# Tri-Learn Graph Fusion Network for Attributed Graph Clustering

**Binxiong Li,[a] Xu Xiang,[b*] Xue Li,[b] Binyu Zhao,[a] Heyang Gao,[a] Qinyu Zhao,[a]**
[a]*Stirling College, Chengdu University, 2025 Chengdu Rd., Chengdu, Sichuan, China, 610106*
[b]*College of Computer Science, Chengdu University, 2025 Chengdu Rd., Chengdu, Sichuan, China, 610106*

E-mail Address:

libinxiong@stu.cdu.edu.cn (Binxiong Li); xiangxu@stu.cdu.edu.cn (Xu Xiang); lixuexue@stu.cdu.edu.cn (Xue Li);

gaoheyang@stu.cdu.edu.cn (Heyang Gao); zhaoqinyu@stu.cdu.edu.cn (Qinyu Zhao);

## ABSTRACT

In recent years, models based on Graph Convolutional Networks (GCN) have made significant strides in the field of graph data analysis. However, challenges such as over-smoothing and over-compression remain when handling large-scale and complex graph datasets, leading to a decline in clustering quality. Although the Graph Transformer architecture has mitigated some of these issues, its performance is still limited when processing heterogeneous graph data. To address these challenges, this study proposes a novel deep clustering framework that comprising GCN, Autoencoder (AE), and Graph Transformer—termed the Tri-Learn Graph Fusion Network (Tri-GFN). This framework enhances the differentiation and consistency of global and local information through a unique tri-learning mechanism and feature fusion enhancement strategy. The framework integrates GCN, AE, and Graph Transformer modules. These components are meticulously fused by a triple-channel enhancement module, which maximizes the use of both node attributes and topological structures, ensuring robust clustering representation. The tri-learning mechanism allows mutual learning among these modules, while the feature fusion strategy enables the model to capture complex relationships, yielding highly discriminative representations for graph clustering. Additionally, the dual self-supervised modules optimize clustering assignments by minimizing KL divergence, ensuring more coherent and accurate cluster differentiation. This comprehensive framework significantly improves clustering performance on attributed graphs. It surpasses many state-of-the-art methods, achieving an accuracy improvement of approximately 0.87% on the ACM dataset, 14.14% on the Reuters dataset, and 7.58% on the USPS dataset. Due to its outstanding performance on the Reuters dataset, Tri-GFN can be applied to automatic news classification, topic retrieval, and related fields.

**The source code for this study is available at https://github.com/YF-W/Tri-GFN**

**Keywords:** Self-supervised learning; Graph Autoencoder; Attribute graph clustering; Graph

Convolutional Neural Networks; Graph Transformer

## 1. Introduction

In recent years, deep learning has made significant breakthroughs in the field of machine learning, providing new insights for clustering tasks [1]. Through the multi-layer nonlinear transformations of deep neural networks, deep learning models can extract multi-level representations of data, capturing complex patterns and features within the data, thus offering more discriminative feature representations for clustering [2]. The fundamental idea of deep learning is to utilize deep neural networks for hierarchical representation of data, thereby decomposing complex patterns into a series of simpler patterns for recognition and processing. Typical deep clustering models include Deep Embedded Clustering (DEC) [3], Deep Clustering Network (DCN) (Yang et al., 2017), and Variational Deep Embedding (VaDE) [4]. These models use autoencoders to map data to a latent space and optimize the distribution of the hidden space by clustering the loss function, thereby achieving effective clustering of the data. However, existing deep clustering methods have challenges in handling high-dimensional data, high computational resource consumption, slow convergence, and dependence on labeled data [5]. In structured deep clustering networks, researchers further combine graph structural information and propose a series of models that fully utilize data structure and meaningful representations. For example, Deep Graph Infomax (DGI) enhances the ability of graph structure representation by maximizing mutual information between global and local representations [6]. Graph Autoencoder (GAE) performs representation learning on graph-structured data through autoencoders, combining clustering losses to achieve accurate segmentation of data [7]. Structural Deep Clustering Network (SDCN) transmits data representations to the corresponding GCN layers via autoencoders and enhances the expression of data structure information [8] through a dual self-supervised mechanism, integrating two different deep neural architectures. These clustering methods utilizing autoencoders and clustering loss effectively extract structural information and useful representations from data, making deep clustering gradually become the mainstream method in the field of clustering, and become an indispensable clustering tool in practical applications.

In the field of graph data analysis, Attributed Graph Clustering (AGC) offers powerful potential for clustering tasks by integrating node attributes and graph structures [9]. For example, the Graph Learning for Attributed Graph Clustering framework achieves more accurate clustering results by jointly modeling graph structures and node attributes [10]. Notably, the GCN proposed by Kipf and Welling achieved significant success in node classification and clustering tasks, sparking a revolution in semi-supervised processing of graph data and demonstrating remarkable effectiveness in node classification and clustering tasks on graph-structured data [11]. Subsequently, Improved Deep Embedded Clustering (IDEC) enhanced the learning ability of node attributes and graph structures by combining incomplete autoencoders with clustering loss functions, reducing the risk of feature space damage and further improving clustering performance [12]. These research advancements collectively propelled the rapid development of the AGC field, providing a reliable foundation for clustering methods that integrate multiple types of information and laying the groundwork for more in-depth future research [13].

Despite the success of GCN in various fields, it still faces challenges in attributed graph clustering, including difficulty in effectively utilizing node attribute information and graph structure information to enhance clustering differentiation and consistency [14,15]; not considering clustering balance as a direct objective or explicitly addressing how to maintain balance, leading to unbalanced allocations; and the inherent over-smoothing problem of GCN [16]. Confronting the current research trends in attributed graph clustering, we find that although previous works like SDCN have made some progress in this area, the problem of effectively bridging the gap between node attributes and graph structure, which are heterogeneous information types, remains unresolved in clustering analysis. On the other hand, Structural Deep Clustering Networks indicate that the effective utilization of structural information in deep clustering is still insufficient. The lack of a mechanism to effectively integrate structural information with data representation has become a key issue in improving the performance of clustering tasks. Clustering methods based on the Transformer architecture have shown good performance in graph data, but they still have some limitations. First, although the multi-head attention mechanism of Transformer models can effectively capture long-distance dependencies between nodes, it lacks targeted optimization in integrating node attributes and graph structure, making it difficult to fully utilize the heterogeneous information between them. Secondly, these models usually require a large number of training parameters, leading to high time and space complexity, making it challenging to apply to large-scale attributed graph data (Zhang et al., 2021). Therefore, further research is needed on how to fully integrate node attributes and graph structure information in attributed graph clustering to better address these challenges.

**Motivations and the limitations of existing works**

In light of the aforementioned work, we find that most existing deep clustering algorithms still face three challenging issues: **(1) the limitation in handling complex data.** Although deep learning models decompose complex patterns by hierarchically representing data, the optimization of latent space is limited when dealing with high-dimensional or complex data, making it difficult to fully deconstruct intricate patterns and achieve optimal clustering. **(2) The challenge of integrating graph structure information with node attributes.** In graph data analysis, effectively combining graph structure information with node attributes remains a significant challenge. **(3) The risk of feature space degradation.** Some methods (such as IDEC) enhance representation capability by integrating autoencoders with clustering loss functions, yet they still face the risk of feature space degradation, potentially impacting the final clustering performance.

**Contributions of the Tri-GFN**

In view of the above research motivations and the limitations of existing works, we propose Tri-GFN. Our model aims to systematically address these limitations through a novel deep neural architecture that leverages GCN, AE, and Graph Transformer. Tri-GFN not only enhances the collaborative representation of node attributes and structural information but also introduces an innovative fusion strategy that boosts clustering differentiation and coherence. This model aims to provide a more comprehensive and robust framework for effectively handling graph data of various sizes and complexities, thereby advancing the theory and practice of attributed graph clustering. Compared to existing methods, the core innovation of Tri-GFN lies in its unique tri-network mutual learning and fusion learning mechanism, meticulously designed to fully exploit the structural and node attribute information of graph data. The model pioneers the aggregation of GCN, AE, and Graph Transformer into a mutually learning network. Through the complementary advantages of these three technologies, Tri-GFN significantly improves the accuracy and consistency of clustering. The contributions are listed as follows.

- **Triple Fusion Learning Mechanism.** The Tri-GFN model achieves a profound integration of node attribute information and graph structure information through its unique architecture. Unlike existing efforts that rely only on a single model or simply combine multiple models, the design of Tri-GFN allows for deep interaction and mutual learning among three distinct mechanisms to reveal and enhance the expressiveness of clustering features.

- **Innovative Application of Self-Attention Mechanism.** Within the context of integrating Graph Transformer technology, Tri-GFN innovatively applies the self-attention mechanism to capture long-distance dependencies in graph data, while effectively mitigating the over-smoothing problem brought by graph convolutional networks. This mechanism enables the model to understand the complex interactions between nodes in a more detailed and comprehensive manner, thereby achieving higher differentiation and consistency in node feature representation for clustering tasks.

- **Triple Channel Enhancement Module.** The triple channel enhancement module designed in Tri-GFN maximizes the use of node attributes and topological information through inter-layer information transmission and structure information integration, achieving comprehensive and robust clustering representation. This module enhances the expression of different information through three complementary paths (GCN path, AE, and Graph Transformer path) and combines node attribute and structure information effectively through a fusion strategy, providing more discriminative representations for clustering tasks.

- **Dual Self-Supervised Modules.** The dual self-supervised modules of Tri-GFN optimize the three network modules towards the same goal through self-supervised strategies, utilizing dual KL divergence losses of soft cluster assignment and target distribution, thereby achieving more robust and consistent clustering representations.

Validated by experiments on seven datasets, Tri-GFN has demonstrated its superiority in clustering quality, as well as its effectiveness and applicability in handling complex real-world network data. The innovations in this study not only advance the development of graph clustering techniques but also provide new ideas and tools for graph-based deep learning research, potentially offering new insights in fields such as social network analysis and bioinformatics.

The structure of this paper is as follows: **Section 1** provides an overview of the importance of attributed graph clustering methods, along with the motivation and innovations of this study. **Section 2** reviews related works on attributed graph clustering and the three core technologies, highlighting their limitations. **Section 3** offers a detailed description of the components and methods of the Tri-GFN model, including a complexity analysis. This section covers symbolic representation, framework structure, GCN, AE, and Graph Transformer modules, as well as the enhancement and self-supervised modules. **Section 4** presents the experimental setup, comparative algorithms, and clustering results of the model, validating its effectiveness through parameter analysis and ablation studies. **Section 5** discusses the potential limitations of the model and analyzes its strengths and weaknesses. **Section 6** concludes the paper by summarizing its contributions and forecasting future research directions.

## 2. Related Work

The emergence of attributed graph clustering has spurred a multitude of innovative approaches, each capable of addressing specific challenges within this domain. This section reviews the evolution from traditional clustering algorithms to modern methods, highlighting their contributions and limitations.

### 2.1 Attribute Graph Clustering

As a commonly used data clustering method, attributed graph clustering has found widespread application in fields such as social network analysis, bioinformatics, recommendation systems, and network security. Early research on traditional clustering algorithms, such as KMEANS [17] and DBSCAN [18], converted graph data into vector representations for clustering analysis. However, these algorithms proved inefficient when handling large-scale graph data and overlooked the graph structure information inherent in the data. Traditional clustering methods performed poorly with large-scale and nonlinear data, leading to the emergence of deep clustering methods that combine deep learning with clustering techniques. These methods leverage deep neural networks to learn data features and representations, enabling more accurate and robust clustering. Common deep clustering methods include autoencoders [19], variational autoencoders [20], generative adversarial networks [21], and graph neural networks [22]. Deep embedding clustering methods map data to a low-dimensional embedding space and employ traditional clustering algorithms within this space. Deep Embedded Clustering (DEC) uses autoencoders to learn sparse data representations and clusters them using the KMEANS algorithm. Deep Clustering (Caron et al., 2018) utilizes deep neural networks for feature learning and integrates iterative clustering algorithms for clustering.

## 2.2 Graph Convolutional Network-based Clustering

With the advancement of deep learning, GCN have demonstrated outstanding performance in handling graph data, becoming an essential tool in the field of graph clustering. GCN and their derivative models have proven significantly effective in graph data analysis, particularly in tasks such as clustering, feature representation learning, and semi-supervised learning. Despite the remarkable progress made by these models in graph data processing, they still face challenges in efficiently integrating graph structural information, accelerating algorithm convergence, and reducing computational resource requirements when dealing with large and complex graph datasets. DFCN [23] combines AE and GCN for deep clustering but needs to address issues of information fusion and scalability. The Adaptive Graph Convolutional Clustering Network [24] improves clustering performance through iterative learning methods but may converge slowly due to complex graph structures and large-scale data. SDCN integrates structural information to enhance clustering effects but faces challenges in effectively utilizing structural information. GCN improve classification accuracy in semi-supervised learning by leveraging node connectivity, yet their scalability is limited. Multi-relational graph embeddings based on GCN address topological redundancy and information aggregation issues, but computational complexity may restrict their application. The Variational Graph Autoencoder (VGAE) proposed by [7], and the subsequent Graph Autoencoder (GAE), use a dual-layer graph convolutional network to learn node representations, excelling in capturing higher-order relationships and global information among nodes.

## 2.3 Attention Mechanism-based Clustering

Despite the excellent performance of GCN in handling graph data, they have limitations in capturing long-distance dependencies, prompting researchers to explore attention mechanism-based methods. The introduction of the attention mechanism offers a new perspective for graph clustering methods. [25] proposed the self-attention model, which improves the performance of the self-attention mechanism by increasing the number of layers or introducing more complex structures. [26] proposed a Self-Attention Graph Pooling (SAGPool) strategy, which selectively aggregates information in the graph through the self-attention mechanism, thus selectively aggregating node features to generate hierarchical representations of the graph, demonstrating the practicality of the self-attention mechanism in graph data. [27] introduced the Graph Attention Network (GAT), which uses the self-attention mechanism to weight the features of neighboring nodes, thereby enhancing node classification efficiency, marking one of the early successful applications of the attention mechanism. [28] proposed the Cross-Attention Enhanced Graph Convolutional Network (CaEGCN), which aims to improve graph clustering by combining the Content Auto-Encoder (CAE) and the GAE modules through a cross-attention fusion mechanism.

## 2.4 Based on Transformer

The introduction of the attention mechanism has ushered in new possibilities, further developing Transformer-based models to better handle complex graph data relationships. The multi-head attention mechanism in Transformer models allows for the parallel computation of multiple attention distributions across different subspaces, thereby enhancing the model's expressive power. A key advantage of the Transformer is its parallelization capability and flexibility concerning sequence length, which enables it to perform well on large-scale datasets. However, despite the significant success of Transformers in NLP, their application to graph-structured data still faces challenges. To overcome these limitations, [29] proposed a novel Graph Transformer model called UGT, designed to enhance the effectiveness of node clustering tasks by maintaining local connectivity and feature-based similarity through the preservation of graph transitivity. Additionally, [30] developed the Transformer-based Dynamic Fusion Clustering Network (TDCN), which employs a dynamic attention mechanism and a special structure-aware operation known as Data Structure Adaptive Transformation. This approach not only effectively integrates the features of Transformer and autoencoder networks but also adapts to the structure and feature information of the input data, significantly improving clustering performance across various datasets.

## 2.5 Limitations

Despite significant progress in attributed graph clustering in recent years, current methods still exhibit limitations in several key aspects. Firstly, many approaches often fail to fully leverage the heterogeneous information between nodes and graph topology to learn discriminative representations. Particularly, when integrating node attributes and graph structural features, they cannot effectively merge these two types of information, thus limiting the potential of the model to fully exploit graph data [31]. In graph data, isolated nodes, i.e., those with very few or no connections to the rest of the graph, may have difficulty obtaining a high-quality embedding representation through the GCN model. This difficulty arises because GCN models rely on the aggregation of local neighborhood information, which isolated nodes lack [32]. Moreover, traditional GCN models are primarily designed for homogeneous graphs and struggle to capture diverse feature information in heterogeneous graph data [33]. Meanwhile, effective fusion of node attributes and structural information is considered to be a key issue in realizing efficient clustering, but this issue has not yet been adequately addressed in the above methods. For instance, the Mutual Boost Network (MBN) model attempts to maintain clustering assignment consistency by enhancing the mutual information between node and structural features [34], yet it still faces challenges in bridging the heterogeneity gap between features. Attention mechanism-based models, such as the GAT model, tend to suffer from over-smoothing when stacked in multiple layers [27]. Although the CaEGCN model combines the strengths of CAE and GAE, effectively merging these types of information without introducing redundancy remains a challenge. The SAGPool model, while selectively aggregating node features, might lose critical information from the graph structure, resulting in incomplete hierarchical representations, and its single pooling strategy may not be able to capture the diversity of features in different graph structures [35]. Even though Transformer-based models provide a robust framework for handling complex data relationships by integrating node features with the graph's topology, they are not specifically optimized for the structural characteristics of graph data.

## 3. Methodology

The methodology outlines the construction and operation of our proposed model. It details the notation, the overall network framework, the specific modules, and the analysis of model complexity.

### 3.1 Notations

An undirected attributed graph can be defined as $\mathcal{G} = \mathcal{V}, \boldsymbol{E}, \mathbf{X}, \mathbf{A}$, where $\mathcal{V} = \{v_1, v_2, \ldots, v_N\}$ represents the set of vertices, and $\mathcal{E}$ denotes the edges between these vertices. The matrix $\mathbf{X} \in \mathbb{R}^{N \times d_0}$ encapsulates the node features, with $\boldsymbol{N}$ indicating the total number of nodes and $d_0$ the dimensionality of each node's feature vector. The adjacency matrix $\mathbf{A} = (a_{ij})_{N \times N} \in \mathbb{R}^{N \times N}$ is the adjacency matrix captures the connectivity between nodes, where $a_{ij} = 1$ if $(v_i, v_j) \in \mathcal{E}$. The normalized adjacency matrix $\widetilde{\mathbf{A}}$ using the formula $\widetilde{\mathbf{A}} = \mathbf{D}^{-\frac{1}{2}}(\mathbf{A} + \mathbf{I})\mathbf{D}^{-\frac{1}{2}}$, where $\mathbf{A} + \mathbf{I}$ represents the adjacency matrix with added self-loops to each node. The degree matrix $\mathbf{D} = diag(d_1, d_2, \ldots, d_N) \in \mathbb{R}^{N \times N}$ is derived from $\mathbf{D}_{\mathbf{ii}} = \sum_j (\mathbf{A} + \mathbf{I})_{ij}$, indicating the degree of each node, including self-loops. All notations are summarized in Table 1.

### 3.2 Overall network framework

In this model, we present a tri-channel graph fusion network structure designed to address graph clustering problems. This architecture primarily consists of the AE module, GCN module, Graph

Transformer module, representation enhancement module, and self-supervised clustering module, shown as Figure 1.

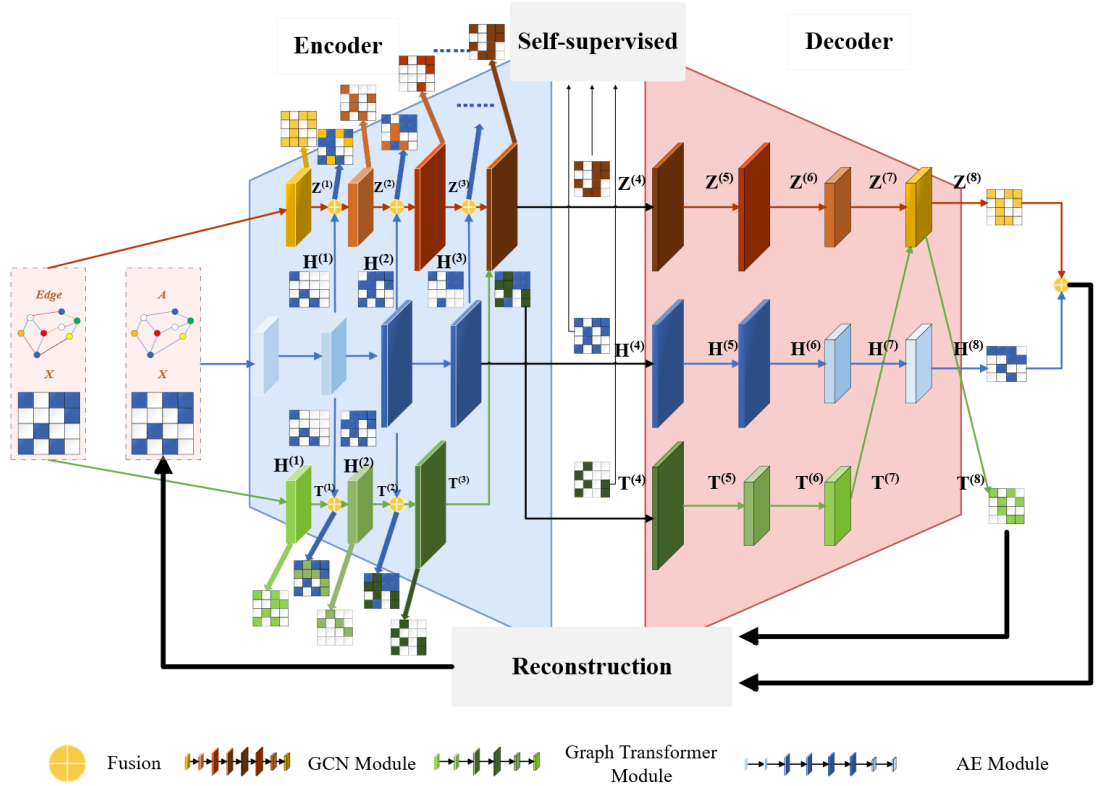

**Fig. 1** The Tri-GFN comprises the following modules: AE, GCN, Graph Transformer module, representation enhancement module, and self-supervised module. In the figure, "Features" and "Edges" refer to the input raw data $\mathbf{X}$ and the edge index matrix $\mathbf{E}$. Initially, we obtain three heterogeneous representations from the graph's attributes and structure through the AE, GCN, and Graph Transformer modules. Here, $\mathbf{Z}^{(4)}$, $\mathbf{H}^{(4)}$ and $\mathbf{T}^{(4)}$ represent the final outputs of the GCN, AE, and Graph Transformer, respectively. After combining these representations, the graph's structural information is further diffused through the adjacency matrix $\mathbf{A}$ to obtain a refined representation. Subsequently, this refined representation is used to compute the soft assignment distribution $\mathbf{Q}$. The clustering soft assignment $\mathbf{Q}'$ is generated from the representation $\mathbf{H}^{(4)}$, and the target distribution $\mathbf{P}$ is derived from $\mathbf{Q}$ to guide the self-supervised clustering process.

### 3.3 AE module

In the Tri-GFN framework, the structure of the AE module is illustrated in Figure 2.

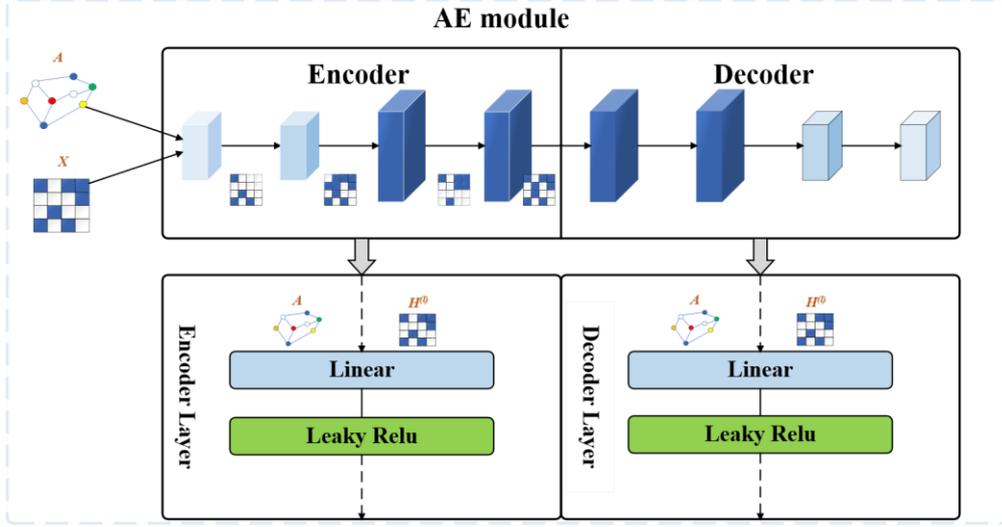

**Fig. 2** AE Module Architecture. The left side illustrates the structure of the input graph, including the vertex feature matrix $\boldsymbol{X}$ and the adjacency matrix $\boldsymbol{A}$. The encoder part consists of multiple layers of linear transformations and Leaky ReLU activation functions, mapping the input features $\boldsymbol{X}$ to latent space representations $\boldsymbol{H}^{(l)}$. The decoder part reverses the operations of the encoder, using the same linear transformations and activation functions to decode the latent space representations $\boldsymbol{H}^{(l)}$ back to the original feature space. The data flow between the encoder and decoder is indicated by arrows. In this manner, the AE module effectively captures the low-dimensional representations of vertex features, achieving feature compression and reconstruction.

A standard auto-encoder employing linear fully-connected layers is implemented to derive representations from node attributes. The representation from a specific $\boldsymbol{\ell}$-th encoder layer of AE can be expressed as:

$$\mathbf{H}^{(\ell+1)} = \sigma\big(\mathbf{W}_e^{(\ell+1)}\mathbf{H}^{(\ell)} + \mathbf{b}_e^{(\ell+1)}\big) \tag{1}$$

In Eq. (1), $l \in \{1,2,\ldots,L-1\}$, where $\sigma(\cdot)$ denotes the activation function, specifically the Leaky Rectified Linear Unit (Leaky ReLU). Additionally, $\mathbf{W}_e^{(\ell+1)}$ and $\mathbf{b}_e^{(\ell+1)}$ signify the weight and bias parameters of the respective encoder layer.

The architecture includes a symmetric decoder that mirrors the encoder, aiming to reconstruct the input data from the latent representation:

$$\hat{\mathbf{H}}^{(\ell+1)} = \sigma\big(\mathbf{W}_d^{(\ell+1)}\hat{\mathbf{H}}^{(\ell)} + \mathbf{b}_d^{(\ell+1)}\big) \tag{2}$$

In Eq. (2), $\mathbf{W}_d^{(\ell+1)}$ and $\mathbf{b}_d^{(\ell+1)}$ are the weight and bias parameters for the corresponding decoder layer. The reconstruction of the input data, $\widehat{\boldsymbol{X}}$ is denoted as $\hat{\mathbf{H}}^{(L)}$. The objective function for data reconstruction is formulated as follows:

$$\mathcal{L}_{res} = \frac{1}{2N} \sum_{i=1}^{N} \| x_i - \hat{x}_i \|_F^2 \qquad (3)$$

In Eq. (3), $x_i \in \mathbf{x}$ and $\hat{x}_i \in \widehat{\mathbf{X}}$ represent the raw feature and reconstructed feature of the same node, respectively. Minimizing the reconstruction error enables the representation to preserve more characteristics of the data, $\mathbf{X}$. Thus, the representation $\mathbf{H} = \mathbf{H}^{(L)}$ is continually refined by extracting node feature information during the training process of the AE module.

## 3.4  GCN Module

In order to comprehensively capture the information in graph-structured data, the structure of the GCN module is illustrated in Figure 3.

We designed a GCN module, which centers on updating the feature representation of nodes using graph convolution operations. By combining the feature information of the nodes and the topology of the graph, the graph convolution network is able to learn the deep relationships and properties of the nodes in the graph. In the following, we describe in detail how the GCN module works and its implementation in our framework. Particularly, the representation learned by the $\ell$ - th layer of the GCN, respectively, $\ell \in \{1, 2, \dots, \frac{L}{2}\}$, acting as a graph encoder, can be formulated through the following convolutional operation:

$$\mathbf{Z}_{GCN}^{(\ell+1)} = \sigma\big(\mathbf{E}\mathbf{Z}_{GCN}^{(\ell)}\mathbf{W}_e^{(\ell)}\big) \qquad (4)$$

In Eq. (5)., the feature matrix $\mathbf{X}$ and the edge index matrix $\boldsymbol{E}$ of the raw data is fed into the first layer of graph encoder to obtain the representation $\mathbf{Z}^{(1)}$. Specifically, $\mathbf{E} = [\text{s}; \text{t}] \in \mathbb{R}^{2 \times \mathcal{E}}$, where s and t are source and target node index vector, encapsulates the graph structure through a set of edge indices, each indicating a connection between two nodes. This compact representation allows the graph encoder to leverage the underlying topology of the data, enabling the aggregation of features from connected nodes according to the graph's adjacency relations. By incorporating both X and E, the graph encoder effectively captures the intrinsic patterns of the graph, facilitating the learning of more informative and representative node embeddings.

For each node in the graph, the updated feature after a graph convolution operation can be calculated by:

$$\mathbf{x}'_i = \Theta^\top \sum_{j \in \mathcal{N}(i) \cup i} \frac{e_{j,i}}{\sqrt{\hat{d}_j \hat{d}_i}} \mathbf{x}_j \tag{5}$$

In Eq. (5)., $\mathbf{x}'_i$ is the updated feature vector of node $\boldsymbol{i}$ after applying the graph convolution. $\boldsymbol{\Theta}$ denotes the weight matrix of the GCN layer, which is learned during the training process. This matrix transforms the aggregated features from the neighborhood into a new feature space. $\boldsymbol{e}_{j,i}$ represents the edge weight from node $\boldsymbol{j}$ to node $\boldsymbol{i}$. If the graph is unweighted, these weights are typically set to 1. For weighted graphs, they reflect the strength or capacity of the connection. Here, $\hat{\boldsymbol{d}}_i = 1 + \sum_{\boldsymbol{j} \in \boldsymbol{\mathcal{N}(i)}} \boldsymbol{e}_{j,i}$ accounts for the degree of node $\boldsymbol{i}$ after adding self-loops (i.e., connections from a node to itself), ensuring that every node considers its own features during aggregation.

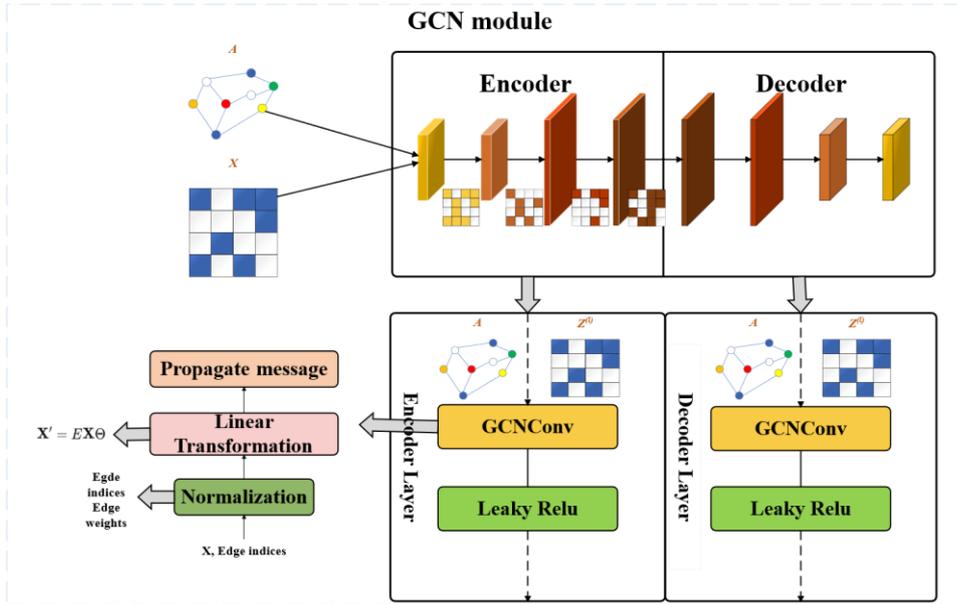

**Fig. 3** The structure of the GCN module is displayed. The top left corner shows the input graph structure, including the vertex features $\boldsymbol{X}$ and the adjacency matrix $\boldsymbol{A}$. The feature matrix $\boldsymbol{X}$ of the input graph is updated to $\boldsymbol{X}'$ through normalization, linear transformation, and information propagation. These feature matrices are processed by the encoder and decoder. The encoder part consists of multiple GCN layers and Leaky ReLU activation functions, while the decoder part performs the inverse operations of the encoder. The data flow between the encoder and decoder is indicated by arrows, with each GCN layer processing $\boldsymbol{A}$ and the updated feature matrix $\boldsymbol{Z}^{(\ell)}$. This architecture is designed to capture complex relationships and feature information among vertices in the graph.

### 3.5 Graph Transformer Module

The Graph Transformer module is designed to process graph data by encoding and decoding node features through a series of Graph Transformer layers, while fusing different feature representations (e.g., the output of a self-encoder) in the process. The module is designed to capture complex relationships between nodes and generate enhanced node embeddings. Each Graph Transformer layer is implemented internally using TransformerConv [36], which allows each node to update itself based on the features of its neighboring nodes. TransformerConv weighs the importance of neighboring nodes through an attention mechanism, allowing the model to focus on key information, thus improving the quality of the representation. The structure of the Graph Transformer module is illustrated in Figure 4.

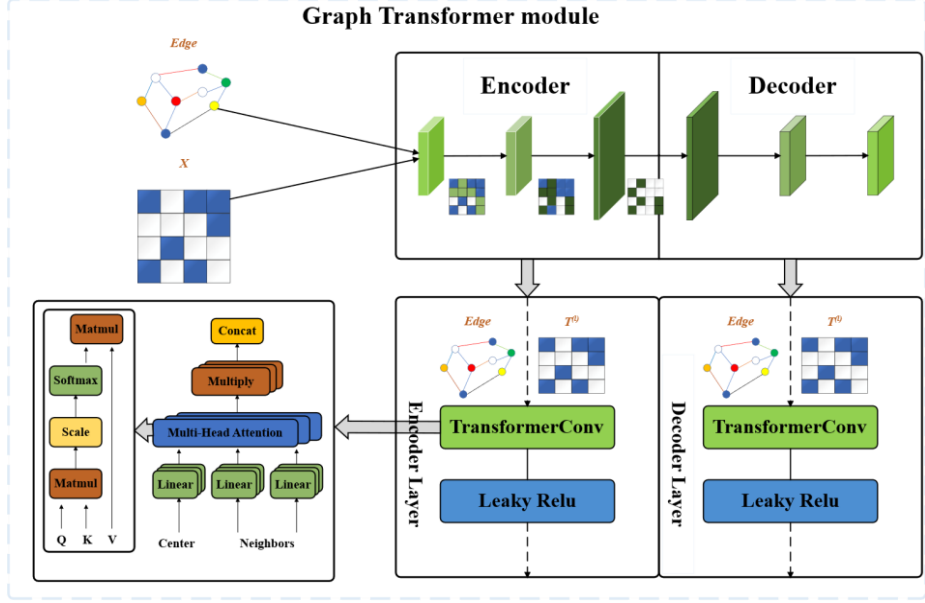

**Fig. 4** Graph Transformer Module Architecture. The top left corner illustrates the input graph structure. The input graph's feature matrix $X$ and adjacency matrix are processed by the encoder and decoder. The encoder consists of multiple TransformerConv layers and Leaky ReLU activation functions, while the decoder performs the inverse operations of the encoder. The data flow between the encoder and decoder is indicated by arrows, with each TransformerConv layer processing the graph structure's edges and the updated feature matrix $T^{(\ell)}$. The bottom left corner details the multi-head self-attention mechanism, including linear transformations, multi-head attention, Softmax operations, and more. This architecture allows the Graph Transformer module to capture complex relationships and feature information between vertices and edges in the graph.

In particular, with the weight matrix $\mathbf{U}_e^{(\ell)}$ of $\ell$-th layer of the graph encoder, respectively, $\ell \in \{1, 2, \ldots, \frac{L}{2}\}$, the representation learned by the $\ell$-th layer of Graph Transformer as graph encoder part, can be obtained by the following self-attention operation:

$$\mathbf{Z}_T^{(\ell+1)} = \sigma\left(\mathbf{E}\mathbf{Z}_T^{(\ell)}\mathbf{U}_e^{(\ell)}\right) \qquad (6)$$

The input features x is first passed through three different linear transformations (corresponding to $\mathbf{Q}$, $\mathbf{K}$, $\mathbf{V}$) to generate keys, queries, and values.

$$\mathbf{Q} = \mathbf{W}_{\text{query}}\mathbf{x}, \mathbf{K} = \mathbf{W}_{\text{key}}\mathbf{x}, \mathbf{V} = \mathbf{W}_{\text{value}}\mathbf{x} \qquad (7)$$

The dot product of the keys and queries is used to compute the attention scores, which are then normalized using a Softmax function, resulting in attention weights for each node with respect to its neighbors.

$$\alpha_{i,j} = \phi\left(\frac{(\mathbf{W}_{\text{query}}\mathbf{x}_i)^{\top}(\mathbf{W}_{\text{key}}\mathbf{x}_j)}{\sqrt{d}}\right) \qquad (8)$$

Where $\phi(\cdot)$ is the nonlinear sigmoid function. $\mathbf{W}_{\text{query}}$ and $\mathbf{W}_{\text{key}}$ are additional learnable weight matrices for transforming the features before computing the dot product. $d$ is the dimensionality of the features, used for normalization to prevent overly large values in the nonlinear sigmoid function. The attention coefficients $\alpha_{i,j}$ are computed using multi-head dot product attention, which allows the model to focus on different parts of the neighbor features. Based on the computed attention weights $\alpha_{i,j}$, the values of each node's neighbors are aggregated, weighted by these attention scores, to form new feature representations for each node:

$$\mathbf{x}'_i = \mathbf{W}_{\text{skip}}\mathbf{x}_i + \sum_{j \in \mathcal{N}(i)} \alpha_{i,j}\mathbf{W}_{\text{value}}\mathbf{x}_j \qquad (9)$$

$\mathbf{x}'_i$ is the updated feature vector of node $i$. It's the new representation of the node after processing through the TransformerConv layer, incorporating information from its own features and the features of its neighboring nodes weighted by attention scores. $\mathbf{W}_{\text{skip}}$ is a weight matrix that is applied to the original features of node $i$ itself. $\mathbf{W}_{\text{value}}$ is a weight matrix transforms the features of the neighboring nodes before they are aggregated.

We assume that the representations obtained through the corresponding $\ell$-th of graph decoder part is defined as:

$$\hat{\mathbf{Z}}_T^{(\ell+1)} = \sigma\left(\mathbf{E}\hat{\mathbf{Z}}_T^{(\ell)}\mathbf{U}_d^{(\ell)}\right) \qquad (10)$$

In Eq. (10)., with the weight matrix $\mathbf{U}_d^{(\ell)}$ of $\ell$-th layer of the graph encoder and decoder, respectively, $\ell \in \{\frac{L}{2}+1, \frac{L}{2}+2, ..., L\}$, the representation learned by the $\ell$-th layer of Graph Transformer as graph decoder part.

### 3.6 Three channel enhancement Module

In our approach, we enhance the utilization of node attributes and the graph's topological structure through a carefully designed Representation Enhancement module. This module operates through two pivotal processes: interlayer information propagation and structural information integration. Initially, to capture more nuanced and discriminative features from the graph data, we seamlessly integrate the nodes' pure attribute features into the GCN for advanced structural representation learning. Subsequently, to effectively propagate these enriched features together with the structural information, we design a method where the latent representations, denoted as $\mathbf{H}^{(\ell)}$, from AE are linearly combined into the respective layers of both the GCN and Graph Transformer modules. The corresponding layer of GCN module and Graph Transformer module by the linear calculation:

$$\mathbf{Z}_{GCN}^{(\ell+1)} = \mathbf{GCN}\left(\varepsilon \mathbf{H}^{(\ell)} + (1-\varepsilon)\mathbf{Z}_{GCN}^{(\ell)}\right) \tag{11}$$

$$\mathbf{Z}_T^{(\ell+1)} = \mathbf{Graph\ Transformer}\left(\varepsilon \mathbf{H}^{(\ell)} + (1-\varepsilon)\mathbf{Z}_T^{(\ell)}\right) \tag{12}$$

Where $\ell \in \{1, 2, \dots, \frac{L}{2}\}$. Here, $\mathbf{GCN}(\cdot)$ indicates the $(\ell+1)-$th graph convolution layer and $\mathbf{Graphormer}(\cdot)$ indicates the $(\ell+1)-$th the self-attention mechanism in graph transformer layer. The coefficient $\varepsilon$ is a balancing factor for the importance of different types of latent features. Consequently, the updated representation $\mathbf{Z}^{(\ell+1)}$ is updated by the introduced operator, and Eq. (5) should be rewritten as:

$$\mathbf{Z}_{GCN}^{(\ell+1)} = \sigma\left(\mathbf{E}\left[\varepsilon \mathbf{H}^{(\ell)} + (1-\varepsilon)\mathbf{Z}_{GCN}^{(\ell)}\right]\mathbf{W}_e^{(\ell)}\right) \tag{13}$$

And Eq. (4) should be rewritten as:

$$\mathbf{Z}_T^{(\ell+1)} = \sigma\left(\mathbf{E}\left[\varepsilon \mathbf{H}^{(\ell)} + (1-\varepsilon)\mathbf{Z}_T^{(\ell)}\right]\boldsymbol{U}_e^{(\ell)}\right) \tag{14}$$

Then, to further enhance the quality of representation, we utilize an operator similar to graph convolutional operation to integrate structural information from neighbors, as shown below:

$$\mathbf{Z}_L = \widetilde{\mathbf{A}}\left(\boldsymbol{\lambda} \cdot \mathbf{Z}_{GCN}^{\left(\frac{L}{2}\right)} + \boldsymbol{\theta} \cdot \mathbf{Z}_{AE}^{\left(\frac{L}{2}\right)} + \boldsymbol{\gamma} \cdot \mathbf{Z}_T^{\left(\frac{L}{2}\right)}\right) \tag{15}$$

Where $\mathbf{Z}^{\left(\frac{L}{2}\right)}$, $\mathbf{H}^{\left(\frac{L}{2}\right)}$ and $\mathbf{Z}_T^{\left(\frac{L}{2}\right)}$ are the final representations of GCN, AE and Graph Transformer. After the structure information diffusion via Eq. (15), the representation can be further enhanced by preserving characteristics on the raw graph. With the help of the RE module, the heterogeneous information is propagated over the dual-network and the model can learn more comprehensive and robust representation for clustering. The coefficients $\boldsymbol{\lambda}$, $\boldsymbol{\theta}$, and $\boldsymbol{\gamma}$ represent different weights. These weights are used to balance the influence of different layers, integrating structural information from neighbors, attribute information, and long-range dependencies, thereby enhancing the quality of representation.

### 3.7 Self-supervised module

The self-supervised module in our framework facilitates training the dual-network by employing pseudo labels, essential for graph clustering, an inherently unsupervised task. This module employs a two-step strategy: (1) calculating the soft clustering assignments $\boldsymbol{Q}$ and $\boldsymbol{Q}'$, and (2) generating the target distribution $\boldsymbol{P}$ from $\boldsymbol{Q}$.

**Soft Clustering Assignment Computation**

Initially, the similarity between each cluster centroid $c_j$ and node representation $z_i$ is computed, adopting the Student's t-distribution as per Van der Maaten & Hinton (2008) [37]. Here, $i$ ranges over all nodes $\{1,2,\dots,N\}$, and $j$ spans all clusters $\{1,2,\dots,k\}$. To initialize cluster centroids, we implement KMEANS with random restarts on representations derived from a pre-trained auto-encoder. In the absence of ground truth labels, we employ cluster sum of squares [38] as an internal validity index to identify the optimal clustering arrangement. The resulting similarities are transformed into a probabilistic distribution using the Student's t-distribution, thus defining the soft clustering assignments $\boldsymbol{Q}$ as follows:

$$q_{ij} = \frac{(1+\| z_i - c_j \|^2/t)^{-\frac{t+1}{2}}}{\sum_{j'} (1+\| z_i - c_{j'} \|^2/t)^{-\frac{t+1}{2}}} \tag{16}$$

**Target Distribution Generation**

Upon establishing the soft clustering assignments $\boldsymbol{Q}$, we focus on refining the data representation by leveraging high-confidence assignments to enhance proximity to cluster centers, thereby increasing cluster cohesion. The target distribution $\boldsymbol{P}$ is computed as follows:

$$p_{ij} = \frac{q_{ij}^2/f_j}{\sum_{j'} q_{ij'}^2/f_{j'}} \tag{17}$$

Where $f_j = \sum_i q_{ij}$ represents the frequency of soft cluster assignments. This normalization process squaring the assignments in $\boldsymbol{Q}$ escalates their confidence, leading to an objective function designed to minimize the KL divergence between $\boldsymbol{Q}$ and $\boldsymbol{P}$:

$$L_{clu} = KL(\boldsymbol{P} \parallel \boldsymbol{Q}) = \sum_i \sum_j p_{ij} log \frac{p_{ij}}{q_{ij}} \tag{18}$$

This loss function helps to optimize the neural network's ability to approximate data representations closer to cluster centers, embodying the essence of self-supervision: $\boldsymbol{P}$ is derived from $\boldsymbol{Q}$ and in turn supervises its update.

**Integrating the GCN, AE and Graph Transformer Module**

The self-supervised module, depicted in Figure 5, integrates feature representations from various sources to enhance model performance. Specifically, the module leverages features generated by AE, a GCN, and additional feature representations. The features $\boldsymbol{Z_{AE}}$ and $\boldsymbol{Z_T}$ undergo a fusion operation, resulting in a combined representation which is subsequently used to generate the distribution $\boldsymbol{Q}$. Concurrently, $\boldsymbol{Z_{AE}}$ directly generates another distribution $\boldsymbol{Q'}$. The module employs Kullback-Leibler (KL) divergence to quantify the dissimilarities between these distributions. The divergence $KL(\boldsymbol{Q} \parallel \boldsymbol{Q'})$ measures the discrepancy between $\boldsymbol{Q}$ and $\boldsymbol{Q'}$, $KL(\boldsymbol{P} \parallel \boldsymbol{Q})$ while compares $\boldsymbol{Q}$ with a predefined standard distribution $\boldsymbol{P}$. The self-supervised loss components, $KL_{loss}$ and $Q'Q_{loss}$, are derived from these divergences. This framework facilitates the refinement of feature representations by aligning them with the inherent structure of the data, thereby improving the model's ability to capture complex patterns through a self-supervised learning approach.

$$L_{con} = KL(\mathbf{Q} \parallel \mathbf{Q'}) = \sum_i \sum_j q_{ij} log \frac{q_{ij}}{q'_{ij}} \tag{19}$$

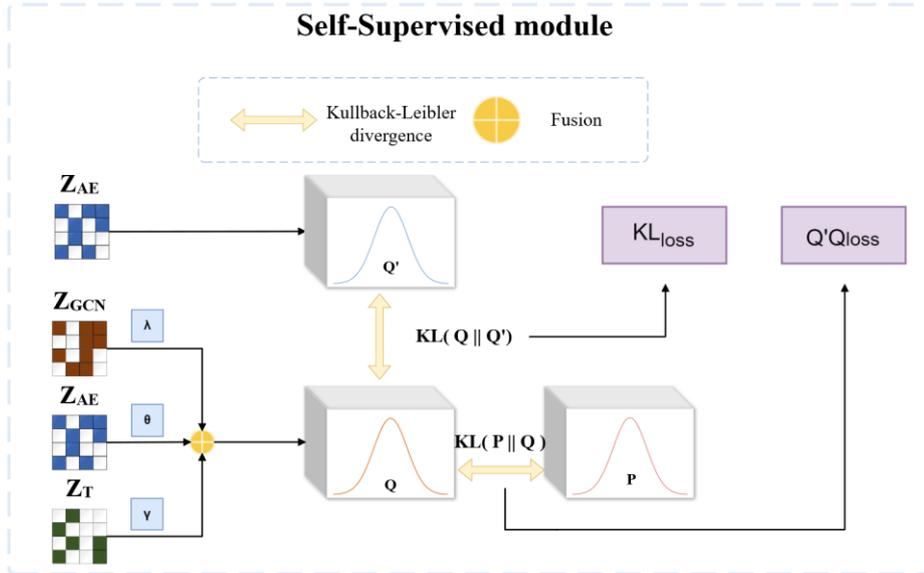

**Fig. 5** The architecture of the self-supervised module is illustrated. The left side shows feature matrices $\boldsymbol{Z_{AE}}$, $\boldsymbol{Z_{GCN}}$ and $\boldsymbol{Z_T}$ obtained from AE, GCN, and Graph Transformer, respectively.

These feature matrices undergo a fusion operation to generate a new distribution $\boldsymbol{Q}$. Then, the KL divergence between $\boldsymbol{Q}$ and the distribution $\boldsymbol{Q}'$ directly generated by the autoencoder is calculated, resulting in the $KL_{loss}$. Simultaneously, the distributions $\boldsymbol{Q}$ and $\boldsymbol{P}$ are compared through KL divergence to optimize the loss function. This approach enables the self-supervised module to learn richer feature representations, enhancing the model's performance.

The objective function of our model is crafted with precision to address complex challenges inherent in graph-based clustering and representation learning. This function not only mitigates common issues associated with traditional learning methods but also strategically leverages the unique capabilities of the GCN, AE, and Graph Transformer modules to optimize performance. Notably, the AE module is instrumental in generating the $\boldsymbol{Q}'$ distribution, distinctively shaping the self-supervision process.

### 3.8 Optimization

Our model employs the optimization function as defined in the GAE module detailed. The loss function used in our framework, represented as Eq. (20), is a composite of two primary components:

$$L_{GFN} = L_w + \delta L_e \tag{20}$$

Here, $L_w$ denotes the loss associated with reconstructing the feature matrix, and $L_e$ pertains to the loss related to reconstructing the adjacency matrix. The balance between these two components is controlled by the parameter $\delta$, which adjusts their relative importance in the overall loss calculation, as further detailed in Eq. (24).

The reconstruction losses are defined as follows:

$$\text{loss}_w = \frac{1}{N} \sum_{i=1}^{N} \left( \widehat{\boldsymbol{Z}_{Gt}} - (\tilde{\mathbf{A}}\mathbf{X})_i \right)^2$$

$$\text{loss}_e = \frac{1}{N^2} \sum_{i=1}^{N} \sum_{j=1}^{N} \left( \widehat{\mathbf{A}}_{ij} - \mathbf{A}_{ij} \right)^2 \tag{21}$$

$$\widehat{\mathbf{A}} = \phi(\hat{\mathbf{Z}}\hat{\mathbf{Z}}^T)$$

$$L_{AE} = \frac{1}{N} \sum_{i=1}^{N} \left( \widehat{\mathbf{H}_{Gt}} - (\tilde{\mathbf{A}}\mathbf{X})_i \right)^2 \tag{22}$$

$$\widehat{\mathbf{H}_{Gt}} = \widehat{\mathbf{H}_{AE}}$$

$$\widehat{\boldsymbol{Z}_G} = \frac{\widehat{\boldsymbol{Z}_{GCN}} + \widehat{\boldsymbol{Z}_T}}{2} \tag{23}$$

Where $\widehat{\boldsymbol{Z}_G} = \frac{\widehat{\boldsymbol{Z}_{GCN}} + \widehat{\boldsymbol{Z}_T}}{2}$ represents the fused graph feature matrix, combining the strengths of both GCN and Graph Transformer outputs. This fusion aims to capture a comprehensive representation of the graph's structural and attribute information. The sigmoid function $\phi(\cdot)$ is employed to transform the product of latent representations into a probability distribution over potential graph edges, facilitating a robust learning framework for node representations without reliance on label guidance. The total loss function for the Tri-GFN model is segmented into three components to cater to the distinct facets of our model:

$$L = L_{rec} + \alpha L_{clu} + \beta L_{con}$$
$$L_{rec} = L_{GFN} + L_{AE} \qquad (24)$$

In this equation, $\alpha$ and $\beta$ are hyperparameters that balance the contributions of the reconstruction loss $L_{rec}$, the clustering loss $L_{clu}$, and the consistency loss $L_{con}$. Optimization of both cluster centers $c_j$ and network parameters for DNN and GAE is conducted using the Adam optimizer (Kingma & Ba, 2017) with added momentum. This is critical for refining the model based on gradients calculated with respect to the clustering loss:

$$\frac{\partial L_{clu}}{\partial c_j} = -\frac{t+1}{t} \sum_{i=1}^{N} (1 + \frac{\| z_i - c_j \|^2}{t})^{-1} \times (p_{ij} - q_{ij})(z_i - c_j) \qquad (25)$$

Upon completion of training, the model achieves convergence in the total training loss. The outputs, including soft assignments $Q$ and $Q'$, guide the selection of $Q'$ as the final clustering result. The prediction label $r$ for each node is determined based on $Q'$ through:

$$\mathbf{r}_i = \arg\max_j q_{ij} \qquad (26)$$

Algorithm 1 shows the procedure of the Tri-GFN.

---

**Algorithm 1:** Tri-Learn Graph Fusion Network for attributed graph clustering

---

**Input**: Graph data (**X**, **Y**, **adj**, **edges**); Hyperparameters $\boldsymbol{\alpha}$, $\boldsymbol{\beta}$, $\boldsymbol{\lambda}$, $\boldsymbol{\theta}$, $\boldsymbol{\gamma}$;

Iteration number $\mathbf{N_{Iter}}$; Cluster number $\mathbf{k}$;

**Output**: Cluster partition results **r**;

---

1    Initialize the parameters of auto-encoder with pre-training;

2    Initialize the weight matrix of GCN;

3    Initialize the weight matrix of Graph Transformer;

4    Initialize $\mathbf{k}$ cluster centroids based on the representation from pre-trained auto-encoder;

5    **for** iter $\in 0, 1, \cdots N_{Iter}$ **do**

6        Obtain the representations $\boldsymbol{Z}^{(l)}$, $\boldsymbol{H}^{(l)}$ and $\boldsymbol{T}^{(l)}$ by Eq. (4), Eq. (1) and Eq. (6);

7        Generate the enhanced representation $\mathbf{Z_L}$ by Eq. (15);

8        Calculate the soft assignments $\mathbf{Q}$, $\mathbf{Q'}$ and $\mathbf{P}$, $\mathbf{Q}$ by Eq. (18) & Eq. (19);

| 9 | Generate the target distribution $\mathbf{P}$ by Eq. (17); |
|---|---|
| 10 | Feed the $\mathbf{Z}$, $\mathbf{H}$ and $\mathbf{T}$ to the decoder to reconstruct the raw data $\mathbf{X}$, $\mathbf{A}$ and $\mathbf{adj}$; |
| 11 | Calculate the objective function by Eq. (24); |
| 12 | Optimize and update parameters of the whole model by back propagation; |
| 13 | **end** |
| 14 | Obtain the results of cluster partition $\mathbf{r}$ by Eq. (26); |

## 3.9 Complexity Analysis

The time complexity of our model primarily stems from the GCN module, AE module, Graph Transformer module, and the self-supervised module. For the GCN module, the computation mainly involves linear transformations and feature aggregation. During the linear transformation, the weight matrix is applied to each node, resulting in a complexity of $O(ND_{in}D_{out})$, where $N$ is the number of nodes, $D_{in}$ is the input feature dimension, and $D_{out}$ is the output feature dimension. In the feature aggregation part, each node collects features from all its neighbors (including itself), leading to a complexity of $O(ED_{out})$, where $E$ is the number of edges in the graph. Assuming that the input and output feature dimensions are approximately the same across all layers, $D_{in} \approx D_{out} \approx D$, the overall time complexity for the GCN module is approximately $O(L(ND^2 + ED))$, where $L$ is the number of layers. The AE typically consists of multiple encoding layers, one or more bottleneck layers, and multiple decoding layers. Assuming the AE module has $L$ layers, including encoding, bottleneck, and decoding layers, the total time complexity is: $O\left(N \cdot \sum_{l=1}^{L} D_{in}^{(l)} \cdot D_{out}^{(l)}\right)$. In the Graph Transformer module, the time complexity of each layer can be considered the time complexity of TransformerConv. The complexity of TransformerConv mainly depends on: $N$: the number of nodes, $D$: the feature dimension of each node, $E$: the number of edges, and $H$: the number of heads in multi-head attention. The time complexity of TransformerConv operations includes: Linear transformation: $O(ND^2)$ per head, resulting in $O(ND^2H)$ in total. Attention computation: $O(ED)$ per head, resulting in $O(EDH)$ in total. Weighted aggregation based on attention scores: $O(EDH)$. Thus, the total time complexity for the Graph Transformer module is: $O(L(N + E) \cdot D^2 \cdot H)$. If the number of layers or the number of attention heads increases, the computational complexity will increase correspondingly, but it remains linearly related to the number of nodes and edges. Notably, if the graph is dense, i.e., the number of edges $E$ approximates $N^2$, the complexity might be more significantly influenced by the square of the number of nodes. For the self-supervised module, the total time complexity is: $O(Nk + NlogN)$, where $k$ is the number of clusters. Overall, the total time complexity of all modules in Tri-GFN is: $O((L(ND^2 + ED)) + \left(N \cdot \sum_{l=1}^{L} D_{in}^{(l)} \cdot D_{out}^{(l)}\right) + (L(N + E) \cdot D^2 \cdot H) + (Nk + NlogN))$.

## 4. Experiments

This section provides a comprehensive evaluation of our model using seven benchmark datasets. It includes detailed descriptions of the datasets, the evaluation criteria used, and comparisons with 13 state-of-the-art methods. Additionally, implementation details and clustering results are provided. We also conducted extensive parameter analysis, including hyperparameter tuning and the impact of different weights, to understand their effects on model performance. Ablation studies examined the contribution of each module in the model. To validate the robustness of the results, statistical significance tests were performed. Finally, we discuss the analysis of the model training process and the impact of different propagation layers on performance.

## 4.1 Benchmark Datasets

We evaluated our model using various types of public datasets, which are: ACM[1], DBLP [39], Citeseer [40], HHAR [41], Cora [42], USPS [43] and Reuters [44]. Detailed descriptions of these seven datasets are as follows:

- ACM: This dataset includes 3025 papers, 5835 authors, and 56 topics, with features derived from keywords in each paper. These papers are from journals such as KDD, SIGMOD, SIGCOMM, MobiCOMM, and VLDB, and can be categorized into three classes: databases, wireless communication, and data mining.

- DBLP: The first version of this dataset contains 629,814 papers and 632,752 citations. Each author's research field is considered a feature. Citation data are extracted from DBLP, ACM, MAG (Microsoft Academic Graph), and other sources. Researchers can use this dataset for tasks such as network clustering, citation impact analysis, identifying influential papers, and topic modeling.

- Citeseer: This dataset includes 3312 scientific publications divided into six classes, with a citation network containing 4732 links. It comprises sparse word bag feature vectors for each document and a list of citation links between documents. Each publication is represented by a binary word vector indicating the presence or absence of 3703 unique words in the dictionary.

- HHAR: This dataset contains 10,299 sensor records used to benchmark human activity recognition algorithms in real-world settings. All samples are classified into six categories. The dataset is collected from smartphones and smartwatches and reflects the sensing heterogeneity expected in real-world deployments through various device models and usage scenarios.

- Cora: This dataset consists of 2708 scientific papers and 5429 edges, divided into seven different categories. Each paper is represented by a 1433-dimensional word vector, thus each sample point contains 1433 features. Each paper in the dataset is described by a binary word vector indicating the presence or absence of words from the corresponding dictionary.

- USPS: USPS is a dataset containing 9298 grayscale handwritten digit images of 16x16

pixels. These images' features are the grayscale values of the pixels, normalized to the range [0, 2]. The dataset comprises ten categories, each corresponding to a digit (0 to 9). Each image is a grayscale image with a size of 16x16 pixels.

- Reuters: Reuters contains approximately 810,000 English news stories labeled with a classification tree structure. Following the SDCN method, we used four root categories: corporate/industrial, government/social, markets, and economics as labels, excluding documents with multiple labels. Due to computational resource constraints, we randomly selected a subset containing 10,000 samples.

*1https://dl.acm.org/*

## 4.2 Evaluation criteria

In order to evaluate in depth the performance of the Tri-GFN model as well as the comparative models, we chose four recognized and commonly used evaluation metrics including Accuracy (ACC), Normalized Mutual Information (NMI) [45], Adjusted Rand Index (ARI) [46], and Macro F1-score. These metrics reflect the performance of the model on different dimensions. Among these four evaluation criteria, higher values represent models with superior performance.

## 4.3 Comparison methods

**KMEANS**. This commonly used clustering algorithm partitions a dataset into a predefined number of clusters, where each data point is assigned to the cluster represented by the nearest mean point.

**AE**. This method is a deep clustering algorithm that utilizes an autoencoder to determine the representation of each sample. These representations then serve as centroids for the KEMANS algorithm, following the standard KMEANS procedure.

**DEC**. This method achieves high-quality clustering by applying a self-encoder for feature learning and data representation, and using a soft assignment function to compute the cluster assignment probability of data points.

**IDEC**. This method improves upon traditional deep embedded clustering by preserving local data structures during the clustering process. This model combines an under-complete autoencoder with a clustering-oriented loss function to simultaneously learn data representations and perform clustering. The integration of these elements allows IDEC to maintain essential data properties, mitigating the risk of feature space corruption and thereby enhancing clustering performance.

**GAE & VGAE**. The methods are unsupervised graph embedding methods that use the GCN layer as an encoder and use a low-dimensional embedding representation of the graph.

**DAEGC** [47]. This method applies the attention mechanism to a graph self-encoder to realize the task of graph clustering and improve the representation learning and clustering performance on graph data.

**ARGA** (Pan et al., 2019). This method is a graph convolutional neural network algorithm that achieves better graph node representation learning and graph generation capabilities by introducing adaptive graph convolution operations, against network and supervised signals, which can better capture the attribute features of the nodes when generating node representations.

**DDGAE** [48]. This method is a novel community detection model that utilizes dual graph attention mechanisms to process multi-view data on graphs, including high-order modularity and attribute information. This model excels in identifying and enhancing community structures within graph data by reconstructing and optimizing node representations through a self-training module. The approach aims to provide a comprehensive method for community detection by effectively integrating diverse data views.

**DCRN** [49]. This method improves the performance of graph data clustering by reducing the information correlation and introducing a propagation regularization term in a dual network structure.

**TDCN**. This method integrates Transformer and AE models to address the challenges in graph-based clustering. This architecture captures both feature and structural data using a dynamic fusion mechanism involving multiple Transformer layers. TDCN is distinguished by its multi-network and multi-scale fusion modules (TDCN-M and TDCN-S) which synergize to enhance node feature extraction and structural data integration. The design aims to provide a more refined data representation conducive to superior clustering outcomes.

**SDCN**. The method simultaneously utilizes a node feature sub-network, a graph structure sub-network, and applies a two-channel network to supervise the clustering while extracting key information at the node features and graph structure, and ultimately obtaining a comprehensive graph representation.

**MBN**. This method is a dual-network deep clustering model designed for graph data, utilizing AE and GAE to process node and structural data simultaneously. This approach enhances clustering performance through mutual learning and interaction between the two networks. A novel representation enhancement mechanism integrates heterogeneous information from both node and structural features to generate high-confidence clustering assignments. Additionally, a consistency constraint ensures alignment between different clustering outputs, further supported by a self-supervised clustering module for joint optimization of clustering and representation learning tasks.

## 4.4 Implementation details

Before training, we pre-trained the autoencoder with a learning rate of 1e−3 for 50 epochs. In the AE, GCN, and Graph Transformer modules, the dimensions of each layer were set to 500, 500, 2000, and $n_z$ respectively. Specifically, $n_z$ as set differently for each dataset, as detailed in Table 1. The representations obtained from the pre-trained autoencoder were clustered using KMEANS, repeating the process 20 times to select the best result. The network parameters were then optimized and updated over different numbers of epochs on the seven benchmark datasets, where an epoch represents a training iteration. To statistically compare the clustering results of our model, we employed a two-sample T-test analysis and compared our model with 13 comparison methods. We used the mean and corresponding standard deviation to evaluate the clustering performance on the seven datasets. A fixed random seed was used to ensure the reproducibility of the experimental results. The learning rates for the different datasets were set according to Table 1. For the seven datasets, the trade-off parameter $\varepsilon$ for merging the representation layers of the AE module with the GCN and Graph Transformer modules was fixed at 0.5.

We varied the trade-off parameters $\lambda, \theta, \gamma$ and $\varepsilon$ for the GCN and Graph Transformer modules in the enhancement module according to different datasets, as detailed in Table 1. For all comparison algorithms, we referenced the results listed in DDGAE, MBN, and SDCN. As TDCN did not experiment on the DBLP and Cora datasets, and DDGAE and DCRN did not experiment on the Reuters, USPS, and HHAR datasets, the corresponding data are not presented in Table 1. The proposed Tri-GFN was implemented on a machine equipped with an Intel i5-1240P CPU, NVIDIA GTX 3070 GPU, and 32 GB of RAM. The deployment environment was on a Windows 10 system running the PyTorch 3.9.0 platform.

**Table 1.** Training parameter configurations and performance evaluations of comparative algorithms for seven benchmark datasets.

| Dataset | epoch | $\alpha$ | $\beta$ | $n_z$ | Learning rate | $\lambda$ | $\theta$ | $\gamma$ | $\varepsilon$ |
|---------|-------|----------|---------|-------|---------------|-----------|----------|----------|---------------|
| ACM     | 200   | 0.12     | 0.1     | 10    | 5e-5          | 0.5       | 0.4      | 0.1      | 0.5           |
| DBLP    | 200   | 0.1      | 0.12    | 10    | 2e-3          | 0.3       | 0.4      | 0.3      | 0.5           |
| Citeseer| 200   | 0.15     | 0.3     | 10    | 4e-5          | 0.3       | 0.5      | 0.2      | 0.3           |
| Cora    | 400   | 0.1      | 0.12    | 10    | 1e-4          | 0.1       | 0.4      | 0.5      | 0.5           |
| HHAR    | 600   | 0.3      | 0.1     | 20    | 1e-4          | 0.1       | 0.4      | 0.5      | 0.5           |
| Reuters | 200   | 0.3      | 0.15    | 20    | 1e-4          | 0.5       | 0.1      | 0.4      | 0.9           |
| USPS    | 400   | 0.1      | 0.1     | 10    | 1e-3          | 0.2       | 0.3      | 0.5      | 0.5           |

## 4.5 Clustering Results

The clustering performance of Tri-GFN and the comparative algorithms is shown in Table 2&3. We provide the following analysis of the experimental results: The results indicate that our model significantly improves clustering accuracy and efficiency by leveraging the relationships between nodes in graph-structured data. Moreover, it demonstrates marked advantages in high-dimensional data, as well as stability and robustness across different datasets. Compared to traditional clustering models, graph neural network-based clustering models, and Transformer-based clustering models, our model shows substantial performance improvements when dealing with high-dimensional, complex data, large-scale datasets, and small-scale datasets. Our research findings reveal that our algorithm excels in all aspects of clustering performance. Firstly, compared to traditional clustering models such as KMEANS and AE, our model outperforms in four metrics across most datasets. When compared to other graph neural network-based clustering models like DEC, IDEC, SDCN, and MBN, our algorithm also shows superior clustering performance. This outcome is attributed to our model's effective utilization of the rich information provided by the graph structure, especially when handling datasets with complex relationships and highly nonlinear structures. This enables the identification of intrinsic patterns and structures within the data, leading to better information retention and more accurate clustering. Secondly, when compared to the Transformer-based clustering model TDCN, although TDCN excels in handling sequential data and capturing long-range dependencies, our model performs comparably or better on most datasets. This is because our model uses the GCN module to extract topological information from graph data, integrates node and structural features using the AE module, and employs the Graph Transformer to capture long-range dependencies, compensating for the limitations of GCN. Furthermore, compared to the aforementioned three types of models, our model also demonstrates superior stability, with less result fluctuation, indicating strong robustness.

**Table 2.** Clustering results on seven datasets (mean ± std) – 1

| Dataset | Metric | K-Means | AE | DEC | IDEC | GAE | VGAE | DAEGC | OURS |
|---------|--------|---------|-----|-----|------|-----|------|-------|------|
| ACM | ACC | 67.31±0.71 | 81.83±0.08 | 84.33±0.76 | 85.12±0.52 | 84.52±1.44 | 84.13±0.22 | 86.94±2.83 | **93.80±0.12** |
| | NMI | 32.44±0.46 | 49.30±0.16 | 54.54±1.51 | 56.61±1.16 | 55.38±1.92 | 53.20±0.52 | 56.18±4.15 | **76.28±0.11** |
| | ARI | 30.60±0.69 | 54.64±0.16 | 60.64±1.87 | 62.16±1.50 | 59.46±3.10 | 57.72±0.67 | 59.35±3.89 | **81.60±0.10** |
| | F1 | 67.57±0.74 | 82.01±0.08 | 84.51±0.74 | 85.11±0.48 | 84.65±1.33 | 84.17±0.23 | 87.07±2.79 | **93.79±0.11** |
| DBLP | ACC | 38.65±0.65 | 51.43±0.35 | 58.16±0.56 | 60.58±0.22 | 61.21±1.22 | 58.59±0.06 | 62.05±0.48 | **78.99±0.10** |
| | NMI | 11.45±0.38 | 25.40±0.16 | 29.51±0.28 | 29.99±0.17 | 30.80±0.91 | 26.92±0.06 | 32.49±0.45 | **47.52±0.13** |
| | ARI | 6.97±0.39 | 30.80±0.91 | 23.92±0.39 | 19.57±0.30 | 22.02±1.40 | 17.92±0.07 | 21.03±0.52 | **51.13±0.17** |
| | F1 | 31.92±0.27 | 26.92±0.06 | 59.38±0.51 | 61.01±0.21 | 61.41±2.23 | 58.69±0.07 | 61.75±0.67 | **78.14±0.12** |
| Citeseer | ACC | 39.32±3.17 | 57.08±0.13 | 55.89±0.20 | 60.49±1.42 | 61.35±0.80 | 60.97±0.36 | 64.54±1.39 | **71.55±0.13** |
| | NMI | 16.94±3.22 | 27.64±0.08 | 28.34±0.30 | 27.17±2.40 | 34.63±0.65 | 32.69±0.27 | 36.41±0.86 | **45.65±0.59** |
| | ARI | 13.43±3.02 | 29.31±0.14 | 28.12±0.36 | 25.70±2.65 | 33.55±1.18 | 33.13±0.53 | 37.78±1.24 | **47.20±0.51** |
| | F1 | 36.08±3.53 | 53.80±0.11 | 52.62±0.17 | 61.62±1.39 | 57.36±0.82 | 57.70±0.49 | 62.20±1.32 | **65.80±0.16** |
| Cora | ACC | 40.25±0.47 | 44.63±0.17 | 40.67±0.26 | 49.00±1.45 | 63.80±1.29 | 64.34±1.34 | 67.21±0.59 | **73.93±0.03** |
| | NMI | 25.08±0.39 | 23.59±0.23 | 18.92±0.29 | 28.83±1.70 | 47.64±0.37 | 48.57±1.09 | 50.63±0.71 | **54.06±0.05** |
| | ARI | 15.35±0.33 | 19.24±0.16 | 12.89±0.42 | 22.09±2.54 | 38.00±1.19 | 40.35±1.46 | 47.33±0.62 | **52.66±0.05** |
| | F1 | 43.71±1.05 | 42.68±0.24 | 29.15±1.30 | 48.85±1.77 | 65.86±0.69 | 64.03±1.34 | 60.82±0.64 | **69.13±0.03** |
| HHAR | ACC | 59.98±0.02 | 68.69±0.31 | 69.39±0.25 | 71.05±0.36 | 62.33±1.01 | 71.30±0.36 | 76.51±2.19 | **84.51±0.34** |
| | NMI | 58.86±0.01 | 71.42±0.97 | 72.91±0.39 | 74.19±0.39 | 55.06±1.39 | 62.95±0.36 | 69.10±2.28 | **80.28±0.61** |
| | ARI | 46.09±0.02 | 60.36±0.88 | 61.25±0.51 | 62.83±0.45 | 42.63±1.63 | 51.47±0.73 | 60.38±2.15 | **73.56±0.61** |
| | F1 | 58.33±0.03 | 66.36±0.34 | 67.29±0.29 | 68.63±0.33 | 62.64±0.97 | 71.55±0.29 | 76.89±2.18 | **84.99±0.39** |
| Reuters | ACC | 54.04±0.01 | 74.90±0.21 | 73.58±0.13 | 75.43±0.14 | 54.40±0.27 | 60.85±0.23 | 65.50±0.13 | **81.86±0.18** |
| | NMI | 41.54±0.51 | 49.69±0.29 | 47.50±0.34 | 50.28±0.17 | 25.92±0.41 | 25.51±0.22 | 30.55±0.29 | **56.10±0.38** |
| | ARI | 27.95±0.38 | 49.55±0.37 | 48.44±0.14 | 51.26±0.21 | 19.61±0.22 | 26.18±0.36 | 31.12±0.18 | **64.41±0.23** |

**Continued Table 2**

| | | ARGA | DDGAE | DCRN | TDCN | SDCN | MBN | OURS |
|---|---|---|---|---|---|---|---|---|
| | F1 | 41.28±2.43 | 60.96±0.22 | 64.25±0.22 | 63.21±0.12 | 43.53±0.42 | 57.14±0.17 | 61.82±0.13 | **73.80±0.11** |
| **USPS** | ACC | 66.82±0.04 | 71.04±0.03 | 73.31±0.17 | 76.22±0.12 | 63.10±0.33 | 56.19±0.72 | 73.55±0.40 | **81.01±0.10** |
| | NMI | 62.63±0.05 | 67.53±0.03 | 70.58±0.25 | 75.56±0.06 | 60.69±0.58 | 51.08±0.37 | 71.12±0.24 | **81.19±0.20** |
| | ARI | 54.55±0.06 | 58.83±0.05 | 63.70±0.27 | 67.86±0.12 | 50.30±0.55 | 40.96±0.59 | 63.33±0.34 | **74.72±0.16** |
| | F1 | 64.78±0.03 | 69.74±0.03 | 71.82±0.21 | 74.63±0.10 | 61.84±0.43 | 53.63±1.05 | 72.45±0.49 | **78.13±0.11** |

**Table 3.** Clustering results on seven datasets (mean ± std) – 2.

| Dataset | Metric | ARGA | DDGAE | DCRN | TDCN | SDCN | MBN | OURS |
|---|---|---|---|---|---|---|---|---|
| **ACM** | ACC | 88.95±0.26 | 92.03±0.18 | 90.51±0.24 | 90.82±0.19 | 90.45±0.18 | 92.99±0.03 | **93.80±0.12** |
| | NMI | 65.33±0.56 | 72.21±0.12 | 68.19±0.31 | 69.2±0.5 | 68.31±0.25 | 74.97±0.08 | **76.28±0.11** |
| | ARI | 69.82±0.67 | 77.87±0.15 | 74.52±0.16 | 74.84±0.46 | 73.91±0.40 | 80.35±0.12 | **81.60±0.10** |
| | F1 | 89.07±0.26 | 92.06±0.13 | 89.95±0.28 | 90.8±0.19 | 90.42±0.19 | 93.00±0.04 | **93.79±0.11** |
| **DBLP** | ACC | 73.90±0.48 | 80.51±0.21 | 76.59±0.32 | - | 66.44±0.16 | 77.82±0.06 | **78.99±0.10** |
| | NMI | 41.35±0.79 | 49.83±0.39 | 44.96±0.28 | - | 38.12±0.1 | 48.30±0.15 | **48.52±0.13** |
| | ARI | 43.65±0.65 | 55.66±0.42 | 47.65±0.39 | - | 39.65±0.11 | 52.57±0.20 | **53.13±0.17** |
| | F1 | 72.91±0.76 | 79.98±0.24 | 76.03±0.40 | - | 65.23±0.15 | 77.31±0.05 | **78.14±0.12** |
| **Citeseer** | ACC | 61.07±0.49 | 71.64±0.24 | 70.86±0.18 | 69.10±0.53 | 65.96±0.31 | 71.39±0.03 | **71.55±0.13** |
| | NMI | 34.40±0.71 | 46.59±0.16 | 45.86±0.35 | 41.61±0.65 | 38.71±0.32 | **46.14±0.07** | 45.65±0.59 |
| | ARI | 34.42±0.70 | 47.87±0.55 | 47.64±0.30 | 43.91±0.56 | 40.17±0.43 | **47.39±0.02** | 47.20±0.51 |
| | F1 | 58.23±0.31 | 65.91±0.23 | 65.83±0.21 | 62.30±0.31 | 63.62±0.24 | 65.61±0.08 | **65.80±0.16** |
| **Cora** | ACC | 64.03±0.71 | 73.76±0.28 | 61.93±0.47 | - | 50.70±0.09 | 72.16±0.04 | **73.93±0.03** |
| | NMI | 44.90±0.36 | **55.49±0.15** | 45.13±1.57 | - | 33.78±0.07 | 54.16±0.17 | 54.06±0.05 |
| | ARI | 35.20±0.44 | 52.54±0.19 | 33.15±0.14 | - | 25.76±0.07 | 50.82±0.15 | **52.66±0.05** |
| | F1 | 61.90±1.27 | **71.39±0.46** | 49.50±0.42 | - | 44.13±0.07 | 63.85±0.04 | 69.13±0.03 |
| **HHAR** | ACC | 63.30±0.80 | - | - | **88.32±0.12** | 84.26±0.17 | 77.84±0.80 | 84.51±0.34 |
| | NMI | 57.10±1.40 | - | - | **82.24±0.40** | 79.90±0.09 | 80.70±0.12 | 80.28±0.61 |
| | ARI | 44.70±1.00 | - | - | **77.22±0.24** | 72.84±0.09 | 71.15±0.13 | 73.56±0.61 |
| | F1 | 61.10±0.90 | - | - | **88.12±0.15** | 82.59±0.08 | 72.55±0.08 | 84.99±0.39 |
| **Reuters** | ACC | 56.20±0.20 | - | - | 81.5±1.09 | 77.15±0.21 | 71.72±0.18 | **81.86±0.18** |
| | NMI | 28.70±0.30 | - | - | 59.28±0.97 | 50.82±0.21 | 52.22±0.18 | **56.10±0.38** |
| | ARI | 24.50±0.40 | - | - | 62.46±1.94 | 55.36±0.37 | 56.60±0.36 | **64.41±0.23** |
| | F1 | 51.10±0.20 | - | - | 66.28±0.23 | 65.48±0.08 | 65.28±0.13 | **73.80±0.11** |
| **USPS** | ACC | 66.80±0.70 | - | - | 80.35±1.07 | 78.08±0.19 | 75.30±0.66 | **81.01±0.10** |
| | NMI | 61.60±0.30 | - | - | 80.39±0.21 | 79.51±0.27 | 76.72±0.13 | **81.19±0.20** |
| | ARI | 51.10±0.60 | - | - | 73.47±1.09 | 71.84±0.24 | 69.42±0.12 | **74.72±0.16** |
| | F1 | 66.10±1.20 | - | - | 77.94±0.36 | 76.98±0.18 | 74.21±0.06 | **78.13±0.11** |

Clustering performance (%) of our method and thirteen baselines.

The **bold** values are the best and the runner-up results.

"-" denotes that the methods do not converge.

When comparing the ACC metric improvement percentage of the Tri-GFN model with the MBN model across different datasets, the results indicate that our model outperforms the MBN model on most datasets, especially on the Reuters dataset. Our model shows performance improvements of approximately 0.87%, 1.50%, 0.22%, 2.54%, 8.57%, and 7.58% on ACM, DBLP, Citeseer, Cora, HHAR, and USPS datasets, respectively. Notably, on the Reuters dataset, our model achieves a significant 14.14% improvement compared to MBN. This highlights our tri-channel model's superior ability to extract critical information from various datasets, providing more accurate clustering results than the dual-channel MBN.

$$\text{Performance Difference \%}_i = \left(\frac{\text{OURS Mean} - \text{Model Mean}_i}{\text{Model Mean}_i}\right) \times 100$$

We then averaged all these individual performance difference percentages to obtain the average performance improvement percentage:

$$\text{Average Performance Increase\%} = \frac{1}{n}\sum_{i=1}^{n}\text{Performance Difference\%}_i$$

This average value provides a comprehensive measure, and the same calculation method was applied for models in other categories. As shown in Table 4, our model consistently outperforms traditional clustering and graph neural network-based models. Specifically, on the DBLP dataset, NMI and ARI metrics show improvements exceeding 200%. For graph neural network-based models, our model achieves 10% to 70% improvements across various datasets and evaluation metrics. These enhancements are attributed to our model's effective feature extraction from graph-structured data. When compared to the Transformer-based TDCN model, our model shows about a 10% improvement in NMI and ARI metrics on the ACM dataset, although certain metrics on the HHAR and Reuters datasets show declines. This may be due to the heterogeneity of the HHAR dataset and the textual complexity and class imbalance of the Reuters dataset, suggesting that further adjustments may be needed for specific scenarios.

**Table 4.** The average percentage performance improvement of our models versus each set of clustered models on the four evaluation metrics.

| Dataset | Metric | Traditional Models | GNN Models | Transformer Model |
|---------|--------|--------------------|------------|-------------------|
| | | Average Performance Increase (%) | Average Performance Increase (%) | Average Performance Increase (%) |
| ACM | ACC | 26.99 | 6.74 | 3.28 |
| | NMI | 94.93 | 23.92 | 10.23 |
| | ARI | 108.00 | 22.51 | 9.03 |
| | F1 | 26.58 | 6.71 | 3.29 |
| DBLP | ACC | 78.98 | 18.59 | - |
| | NMI | 207.39 | 36.37 | - |
| | ARI | 367.38 | 84.77 | - |
| | F1 | 167.53 | 17.51 | - |
| Citeseer | ACC | 53.66 | 11.79 | 3.55 |
| | NMI | 117.32 | 27.30 | 9.71 |
| | ARI | 156.24 | 31.15 | 7.49 |

| | | | | **Continued Table 3** |
|---|---|---|---|---|
| | F1 | 52.34 | 8.29 | 5.62 |
| Cora | ACC | 74.81 | 25.78 | - |
| | NMI | 122.69 | 40.43 | - |
| | ARI | 201.82 | 70.69 | - |
| | F1 | 61.40 | 32.94 | - |
| HHAR | ACC | 31.96 | 18.46 | -4.31 |
| | NMI | 24.40 | 18.55 | -2.38 |
| | ARI | 40.73 | 30.43 | -4.74 |
| | F1 | 36.89 | 21.79 | -3.55 |
| Reuters | ACC | 30.39 | 24.46 | 0.44 |
| | NMI | 23.98 | 57.87 | -5.36 |
| | ARI | 80.22 | 91.64 | 3.12 |
| | F1 | 49.92 | 27.48 | 11.35 |
| USPS | ACC | 17.64 | 16.51 | 0.82 |
| | NMI | 24.93 | 21.14 | 1.00 |
| | ARI | 31.99 | 29.28 | 1.70 |
| | F1 | 16.32 | 14.79 | 0.24 |

"-" denotes that the methods do not converge.

## 4.6 Parameter Analysis

Our parameter analysis systematically evaluated the impact of different hyperparameter combinations on algorithm performance across multiple datasets. This included assessing weight parameters, fusion coefficients, and loss function weights. The importance and influence of these hyperparameters were validated through comprehensive metrics.

### 4.6.1 Module Weight Hyperparameter Analysis

In a multi-parameter system, understanding the specific impact of different hyperparameters such as $\lambda$, $\theta$, and $\gamma$ on algorithm performance is crucial. We analyze the influence of these hyperparameters through a comprehensive metric—an average of four key performance indicators (ACC, NMI, ARI, F1). Specifically, we designed $\lambda + \theta + \gamma = 1$. In our study, we calculated the comprehensive metric from seven different datasets and generated heatmaps for parameters $\lambda$ and $\theta$ for each dataset, as shown in Figure 6. This step visually demonstrates how these two parameters affect performance, providing foundational data for subsequent analysis. Finally, we constructed a predictive model based on random forests [50] to evaluate the performance of the comprehensive metric under different parameter combinations. The experiments indicate that $\lambda$, $\theta$, and $\gamma$ have almost equal influence on the comprehensive metric, and the overall prediction accuracy of the model is high, validating the effectiveness of the analysis.

**Impact of Parameter Variations on the Comprehensive Metric**

For different datasets, we observed that the comprehensive metric reached its highest value under specific parameter configurations. The optimal parameter settings for each dataset are shown in Table 5. The experimental results reflect the unique attribute configurations of different datasets, allowing the model to maximize its performance under specific parameters. In this study, we systematically adjusted the weight parameters $\lambda$, $\theta$, and $\gamma$ for GCN, AE, and Graph Transformer respectively, to explore the impact of different parameter configurations on the comprehensive metric for each dataset.

We found that for datasets with simple structures and intuitive relationships like ACM, a higher GCN weight ($\lambda$) is crucial, indicating that direct extraction of structural information significantly enhances clustering performance. Conversely, for datasets with complex patterns or dependencies such as HHAR and Cora, the weight of the Graph Transformer ($\gamma$) is particularly important, reflecting the importance of capturing deep relationships and long-distance dependencies in understanding these data structures. Additionally, for datasets rich in attribute information like Citeseer, a higher weight for the AE ($\theta$) demonstrated the significant role of node attributes in the clustering process.

**Table 5.** The highest composite indices and their corresponding parameter sets across seven datasets.

| Dataset | $\lambda$ | $\theta$ | $\gamma$ | The highest composite index |
|---------|-----------|----------|----------|------------------------------|
| ACM | 0.5 | 0.4 | 0.1 | 0.8694 |
| DBLP | 0.3 | 0.4 | 0.3 | 0.6499 |
| Citeseer | 0.3 | 0.5 | 0.2 | 0.5788 |
| HHAR | 0.1 | 0.4 | 0.5 | 0.8182 |
| Reuters | 0.5 | 0.1 | 0.4 | 0.6945 |
| Cora | 0.1 | 0.4 | 0.5 | 0.6233 |
| USPS | 0.2 | 0.3 | 0.5 | 0.7838 |

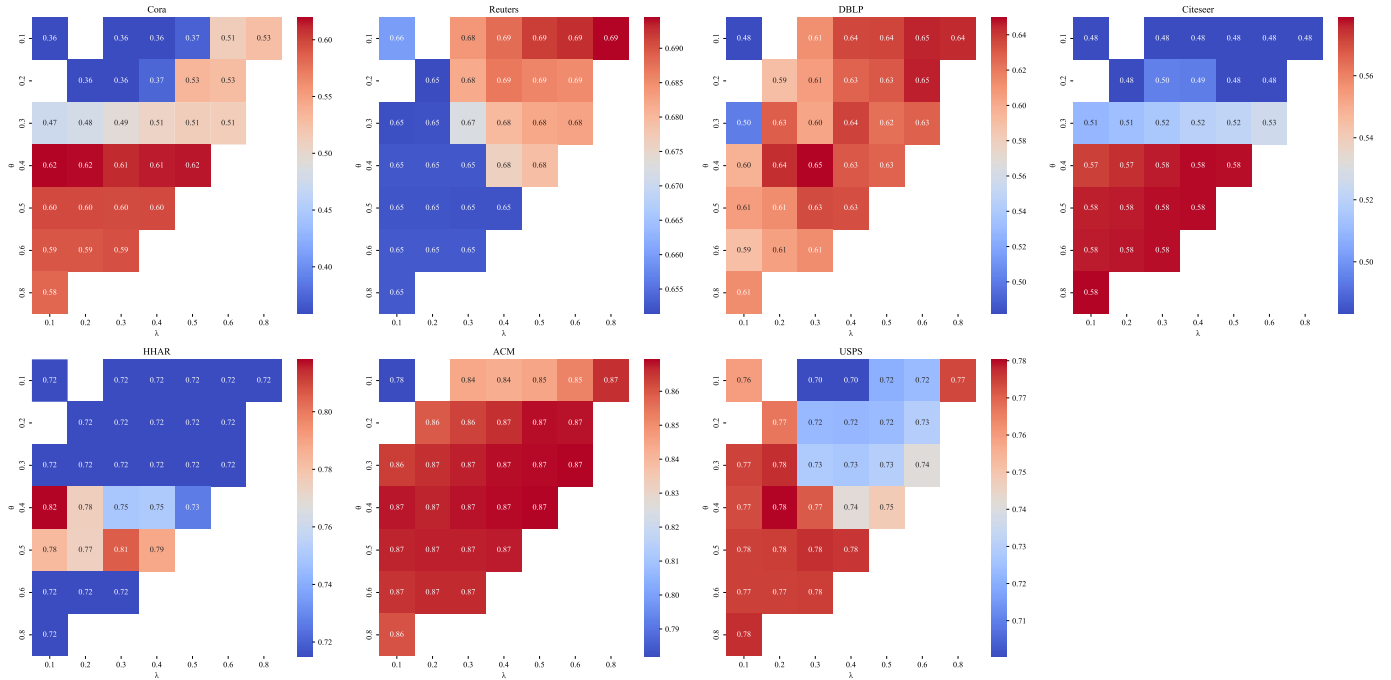

**Fig. 6** Heatmap of the seven datasets, illustrating the impact of $\lambda$, $\theta$ and $\gamma$ on the composite metrics.

**Feature Importance of Weight Hyperparameters**

Our feature importance analysis using a random forest model indicates that $\lambda$, $\theta$, and $\gamma$ have almost equal influence on the composite index, with the model demonstrating high overall predictive accuracy, thereby validating the effectiveness of the analysis. Additionally, we evaluated the random forest model's performance using the MSE method. The MSE was 0.021, a relatively low value, indicating a small discrepancy between the model's predictions and the actual values on the test set. In the random forest model, we calculated the feature importance of each parameter by observing the contribution of individual features at split points across multiple decision trees. The results were as follows: $\lambda$: 29.35%, $\theta$: 35.52%, $\gamma$: 35.13%. These values suggest that $\theta$ and $\gamma$ have nearly identical importance, slightly higher than $\lambda$, in predicting the composite index. This indicates that $\theta$ and $\gamma$ are the key parameters significantly impacting the model's predictive output. Although $\lambda$ is also an important feature, its influence is slightly less compared to the other two parameters.

**Analysis of the $\varepsilon$ Parameter**

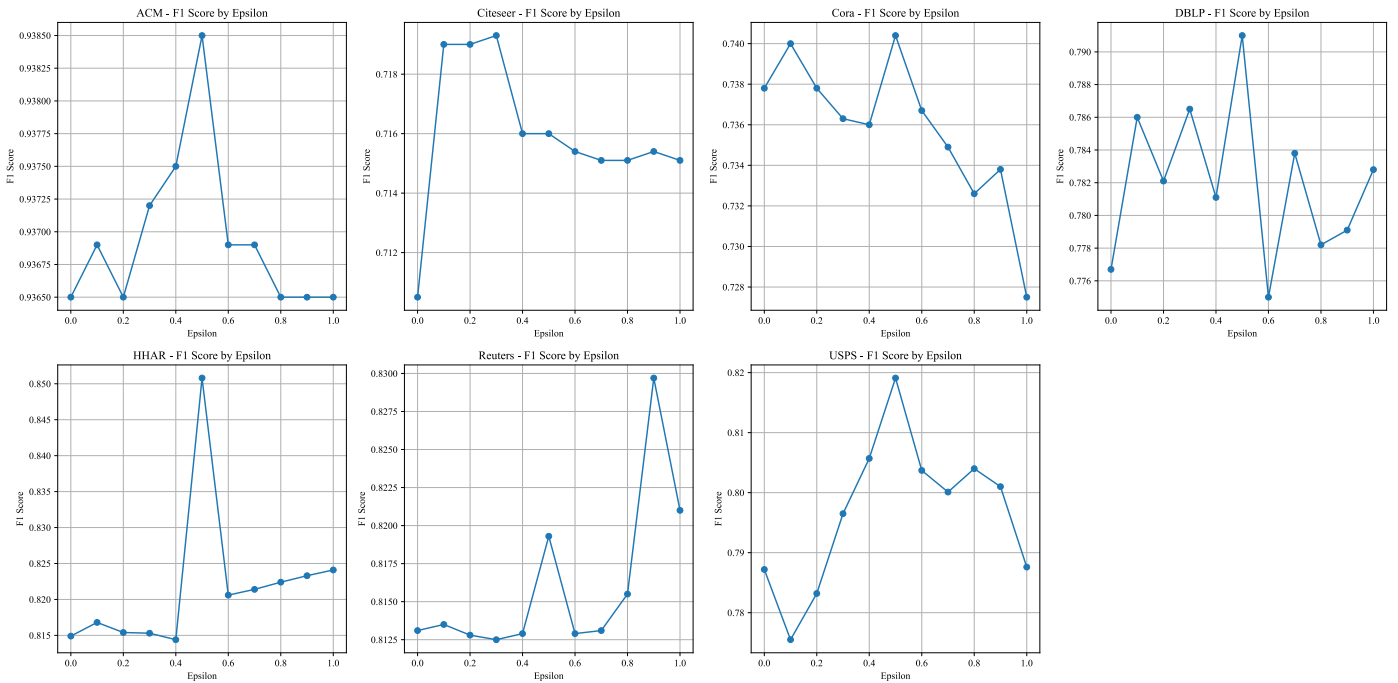

**Fig. 7** Curves of Tri-GFN on seven datasets for F1 score variation with $\varepsilon$.

The $\varepsilon$ parameter is a network fusion coefficient that controls the weight of the feature expressions of each layer of the GCN module and the AE module, and the weight of the feature expressions of each layer of the Graph Transformer module and the AE module in the fusion process. To investigate the impact of $\varepsilon$ on the model's clustering results, we adjusted $\varepsilon$ within the range [0.0, 1.0] in increments of 0.1. The experimental results are shown in Figure 7. From the figure, it can be observed that for most datasets, the F1 score reaches its maximum when $\varepsilon$ increases to 0.5. At this point, the GCN module, AE module, and Graph Transformer module achieve optimal fusion. Although our model performs best on the Citeseer and Reuters datasets when $\varepsilon$ is set to 0.3 and 0.9 respectively, we opted to fix $\varepsilon$ at 0.5 for all datasets. This choice of $\varepsilon$ = 0.5 ensures a balance and consistency in performance across all datasets.

### 4.6.2 Hyperparameter Analysis of the Total Loss Function Weights

To evaluate the impact of the parameters $\alpha$ and $\beta$ within the total loss function on the model, we conducted a hyperparameter analysis. Here, $\alpha$ and $\beta$ represent the weights of the self-supervised clustering loss and the consistency constraint loss, respectively. We selected values for $\alpha$ and $\beta$ from $\{0.01, 0.05, 0.1, 0.12, 0.15, 0.3\}$ to explore their influence on the model. Figure 8 visualizes the experimental results as F1 scores across seven datasets, showing that variations in $\alpha$ and $\beta$ significantly affect clustering performance. Despite some local variations, the results are generally stable overall. The optimal values for $\alpha$ and $\beta$ vary across different datasets and metrics, which are listed in Table 6. Dataset characteristics, such as data distribution, inherent structure, and noise level, influence the optimal weights. For instance, for the structurally clear ACM dataset, lower weights for $\alpha$ and $\beta$ suffice, while for complex or class-imbalanced datasets like Citeseer, higher weights help the model focus on subtle structural differences and consistency.

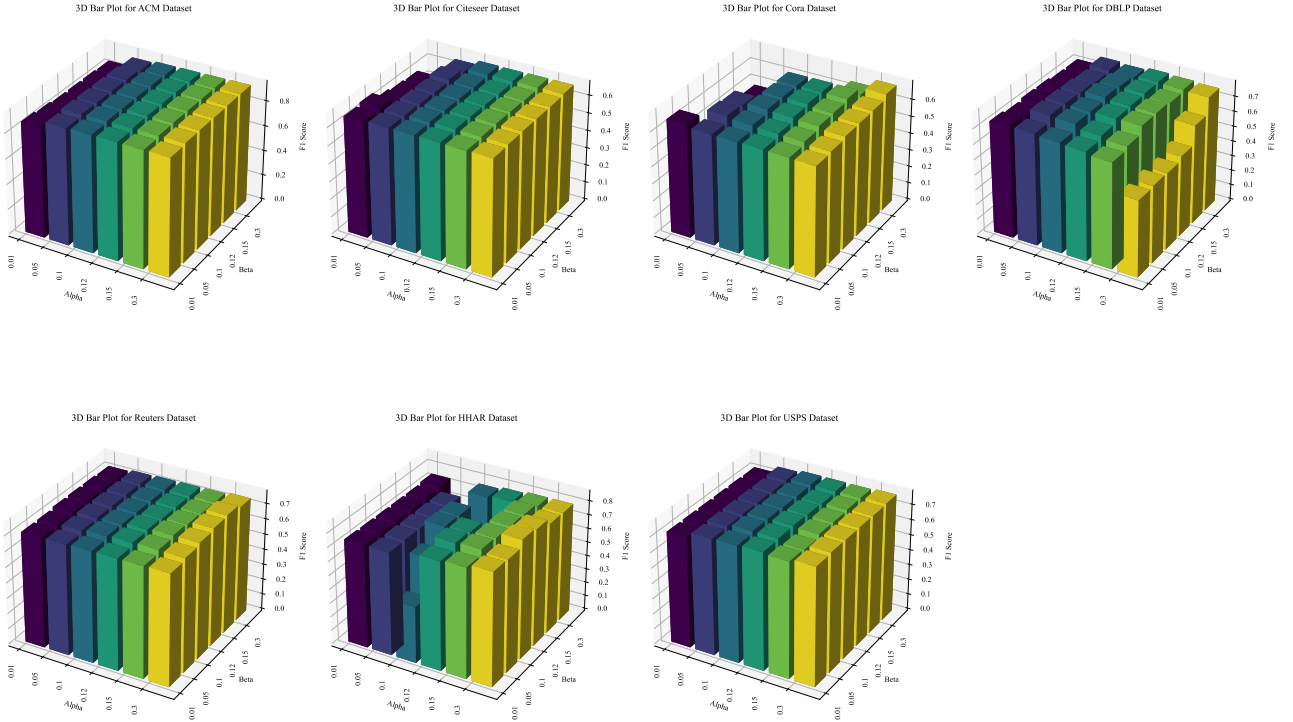

**Fig. 8** A 3D bar chart of the seven datasets, illustrating the variation in F1 scores with changes in the parameters $\alpha$ and $\beta$.

**Table 6.** The optimal $\alpha$ and $\beta$ settings and corresponding performance values for each dataset under different performance indicators (ACC, NMI, ARI, F1).

| Dataset | Best ACC | Best ACC (Alpha, Beta) | Best NMI | Best NMI (Alpha, Beta) | Best ARI | Best ARI (Alpha, Beta) | Best F1 | Best F1 (Alpha, Beta) |
|---|---|---|---|---|---|---|---|---|
| ACM | 0.9385 | (0.12, 0.1) | 0.7739 | (0.12, 0.1) | 0.826 | (0.12, 0.1) | 0.9385 | (0.12, 0.1) |
| Citeseer | 0.7181 | (0.1, 0.3) | 0.4676 | (0.15, 0.3) | 0.4773 | (0.1, 0.15) | 0.6673 | (0.1, 0.15) |
| Cora | 0.7404 | (0.1, 0.12) | 0.5424 | (0.1, 0.12) | 0.5181 | (0.1, 0.12) | 0.6937 | (0.1, 0.12) |
| DBLP | 0.791 | (0.1, 0.12) | 0.4864 | (0.1, 0.12) | 0.5359 | (0.1, 0.12) | 0.7864 | (0.1, 0.12) |
| Reuters | 0.828 | (0.3, 0.15) | 0.5782 | (0.3, 0.15) | 0.6626 | (0.3, 0.3) | 0.7681 | (0.3, 0.15) |
| HHAR | 0.8508 | (0.3, 0.1) | 0.8171 | (0.3, 0.1) | 0.748 | (0.3, 0.1) | 0.8569 | (0.3, 0.1) |
| USPS | 0.8191 | (0.1, 0.1) | 0.8179 | (0.3, 0.05) | 0.747 | (0.1, 0.1) | 0.7745 | (0.1, 0.1) |

## 4.7 Ablation Analysis

In ablation experiments on seven datasets, we compared the full model ("Norm") with five configurations, focusing on module and reconstruction ablation. We analyzed the average performance across four metrics to validate each component's contribution. The results are shown in Table 7&8.

**Table 7.** Module ablation analysis results on seven datasets under four metrics.

| Operation / Dataset | ACC | | | | | | NMI | | | | | |
|---|---|---|---|---|---|---|---|---|---|---|---|---|
| | norm | -AE | -GCN | -T | -dec_GCN | -dec_T | norm | -AE | -GCN | -T | -dec_GCN | -dec_T |
| ACM | 0.9385 | 0.9021 | 0.9372 | 0.9385 | 0.9385 | 0.9388 | 0.7744 | 0.6679 | 0.7714 | 0.7740 | 0.7739 | 0.7742 |
| Citeseer | 0.7169 | 0.6237 | 0.7148 | 0.7157 | 0.7157 | 0.7154 | 0.4676 | 0.3550 | 0.4586 | 0.4582 | 0.4582 | 0.4580 |
| Cora | 0.7404 | 0.4815 | 0.7315 | 0.7334 | 0.733 | 0.7341 | 0.5424 | 0.2731 | 0.5325 | 0.5358 | 0.5349 | 0.5364 |
| DBLP | 0.7910 | 0.7262 | 0.7515 | 0.7816 | 0.7774 | 0.7424 | 0.4864 | 0.4126 | 0.4372 | 0.4757 | 0.4730 | 0.4143 |
| HHAR | 0.8508 | 0.7564 | 0.8403 | 0.8124 | 0.7564 | 0.8077 | 0.8171 | 0.7167 | 0.7998 | 0.7669 | 0.7167 | 0.7817 |
| Reuters | 0.8280 | 0.8295 | 0.7995 | 0.8189 | 0.8183 | 0.8129 | 0.5782 | 0.5672 | 0.5285 | 0.5611 | 0.5598 | 0.5614 |
| USPS | 0.8191 | 0.7650 | 0.7804 | 0.8076 | 0.8166 | 0.781 | 0.7969 | 0.7907 | 0.8119 | 0.7924 | 0.7923 | 0.8112 |

| Operation / Dataset | ARI | | | | | | F1 | | | | | |
|---|---|---|---|---|---|---|---|---|---|---|---|---|
| | norm | -AE | -GCN | -T | -dec_GCN | -dec_T | norm | -AE | -GCN | -T | -dec_GCN | -dec_T |
| ACM | 0.8262 | 0.7286 | 0.8228 | 0.8252 | 0.8260 | 0.8259 | 0.9385 | 0.9027 | 0.9371 | 0.9385 | 0.9385 | 0.9388 |
| Citeseer | 0.4724 | 0.3499 | 0.473 | 0.4747 | 0.4747 | 0.4743 | 0.6673 | 0.6002 | 0.6572 | 0.6581 | 0.6581 | 0.6578 |
| Cora | 0.5181 | 0.1999 | 0.5071 | 0.5072 | 0.5071 | 0.5117 | 0.6924 | 0.4793 | 0.6891 | 0.6922 | 0.6915 | 0.6916 |
| DBLP | 0.5359 | 0.4197 | 0.4774 | 0.5218 | 0.5167 | 0.4506 | 0.7864 | 0.7203 | 0.7397 | 0.7761 | 0.7719 | 0.7388 |
| HHAR | 0.7480 | 0.6422 | 0.7297 | 0.6817 | 0.6422 | 0.7045 | 0.8569 | 0.7452 | 0.8453 | 0.8164 | 0.7452 | 0.8122 |
| Reuters | 0.6567 | 0.6545 | 0.5977 | 0.6448 | 0.6436 | 0.6395 | 0.7681 | 0.7393 | 0.7124 | 0.7376 | 0.7364 | 0.7426 |
| USPS | 0.7470 | 0.7118 | 0.7377 | 0.7268 | 0.7323 | 0.7346 | 0.7720 | 0.7548 | 0.7704 | 0.7720 | 0.7733 | 0.7678 |

norm: Indicates the results obtained with the normal, unmodified model.

-AE: Shows the model's performance after the removal of the AE component.

-GCN: Represents the results when the GCN module is omitted.

-T: Depicts the performance metrics without the Graph Transformer module.

-dec_GCN: The results after the removal of the decomposed Graph Convolutional Network.

-dec_T: Shows the impact on model performance when the decomposed Graph Transformer module is excluded.

The **bold** values are the best results and the <u>Underlined values</u> are the runner-up results.

For module ablation, we removed the AE module, GCN module, Graph Transformer component, the GCN decoder (dec_GCN), and the Graph Transformer decoder (dec_T) from the Norm model. The complete model performed best, achieving the highest scores across all metrics. Removing the AE module significantly reduced performance, with average accuracy dropping to 0.726, NMI to 0.540, ARI to 0.530, and F1 score to 0.706. Removing the GCN and dec_GCN had smaller but still notable impacts on ARI and NMI. The Graph Transformer and its decoder also had a significant role, as their removal slightly decreased average accuracy and other metrics. These results highlight the importance of the AE and Graph Transformer components.

For reconstruction ablation, we assessed the impact of graph reconstruction information from the GCN, Graph Transformer, and their combination on graph reconstruction loss. The GCN performed best in this task, achieving an average ACC score of 0.8107, compared to 0.7758 for the Graph Transformer and 0.7965 for the combined GCN-Graph Transformer. These findings underscore the GCN's superiority in graph reconstruction.

**Table 8.** Reconstruction ablation analysis results on seven datasets under four metrics.

| Metrics | ACC | | | NMI | | |
|---|---|---|---|---|---|---|
| Operation<br>Dataset | GCN | Graph<br>Transformer | GCN-<br>Graph<br>Transformer | GCN | Graph<br>Transformer | GCN-<br>Graph<br>Transformer |
| ACM | **0.9385** | <u>0.9372</u> | <u>0.9372</u> | **0.7744** | <u>0.7705</u> | 0.7698 |
| Citeseer | **0.7160** | <u>0.7166</u> | 0.7160 | <u>0.4605</u> | 0.4653 | **0.4665** |
| Cora | **0.7404** | 0.7352 | <u>0.7363</u> | **0.5424** | <u>0.5401</u> | 0.5334 |
| DBLP | **0.7910** | 0.6288 | <u>0.7764</u> | **0.4864** | 0.3456 | <u>0.4658</u> |
| HHAR | **0.8508** | 0.8136 | <u>0.8139</u> | **0.8171** | 0.7866 | <u>0.7912</u> |
| Reuters | **0.8193** | <u>0.8127</u> | 0.8120 | **0.5626** | 0.5572 | <u>0.5574</u> |
| USPS | **0.8191** | <u>0.7866</u> | 0.7834 | **0.7969** | 0.8117 | **0.8142** |
| Metrics | ARI | | | F1 | | |
| Operation<br>Dataset | GCN | Graph<br>Transformer | GCN-<br>Graph<br>Transformer | GCN | Graph<br>Transformer | GCN-<br>Graph<br>Transformer |
| ACM | **0.8262** | <u>0.8225</u> | <u>0.8225</u> | **0.9385** | <u>0.9372</u> | 0.9371 |

|  |  |  |  |  |  | Continued Table 8 |
|---|---|---|---|---|---|---|
| **Citeseer** | <u>0.4757</u> | 0.4730 | **0.4776** | <u>0.6589</u> | **0.6632** | 0.6531 |
| **Cora** | <u>0.5181</u> | **0.5197** | 0.5140 | **0.6924** | 0.6845 | <u>0.6923</u> |
| **DBLP** | **0.5359** | 0.3204 | **0.5126** | **0.7864** | 0.6043 | <u>0.7706</u> |
| **HHAR** | **0.7480** | 0.7103 | <u>0.7147</u> | **0.8569** | <u>0.8181</u> | 0.8180 |
| **Reuters** | **0.6456** | <u>0.6326</u> | 0.6321 | <u>0.7392</u> | **0.7396** | 0.7323 |
| **USPS** | **0.7470** | <u>0.7418</u> | 0.7404 | <u>0.7720</u> | 0.7758 | **0.7744** |

GCN: Graph reconstruction using GCN represents the result of calculating graph reconstruction loss.

Graph Transformer: Graph reconstruction using Graph Transformer represents the result of calculating graph reconstruction loss.

GCN-Graph Transformer: The average graph reconstruction of GCN and Graph Transformer is used to represent the result of calculating the graph reconstruction loss.

The **bold** values are the best results and the <u>Underlined values</u> are the runner-up results.

## 4.8 Significance Experiments

### Experimental Design and Methodology

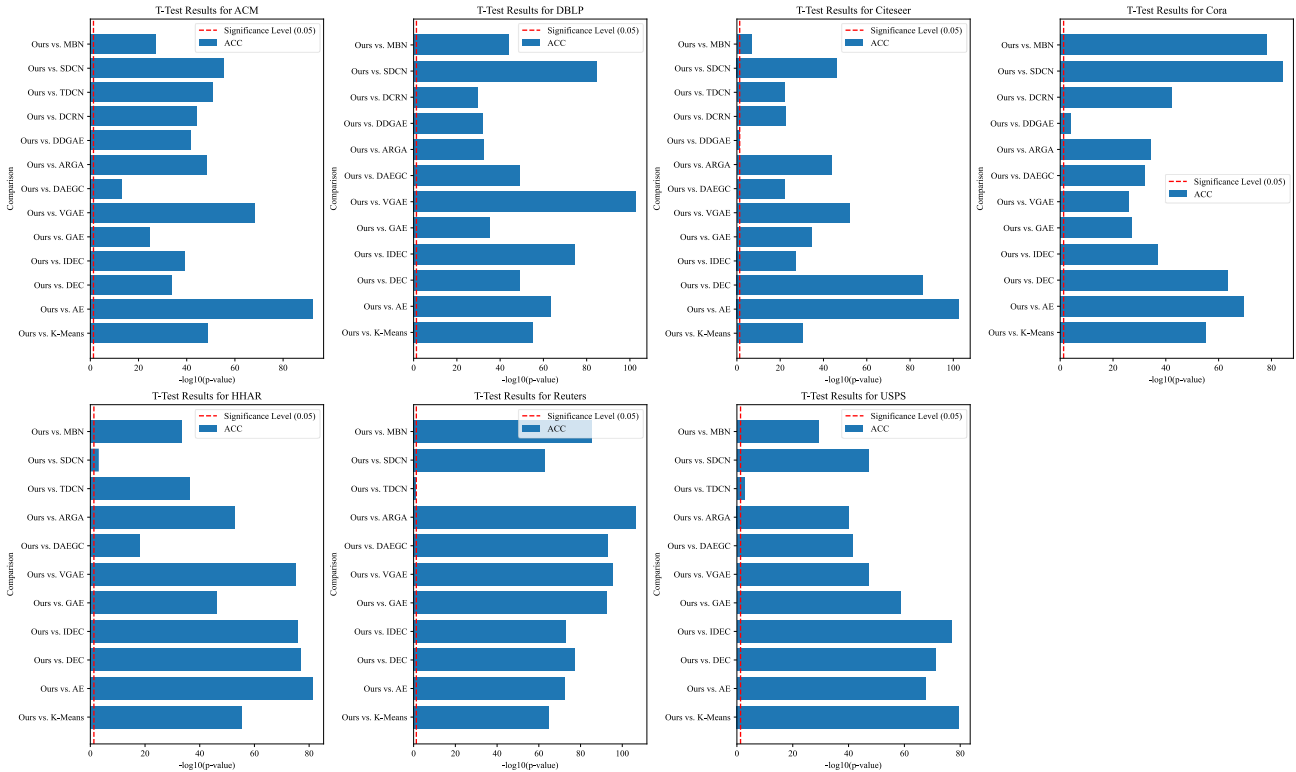

**Fig. 9** The charts present the results of two-sample t-tests comparing our method (Ours) against various baseline algorithms (KMEANS, AE, DEC, IDEC, GAE, VGAE, DAEGC, ARGA, DDGAE, DCRN, TDCN, SDCN and MBN) across different datasets (ACM, DBLP, Citeseer, Cora, HHAR, Reuters and USPS). Each bar represents a comparison against a different method (for instance, Ours

vs. AE), with the x-axis indicating the negative log10 p-value (-log10(p-value)) to demonstrate the extent of statistical significance. The red vertical line represents the level of statistical significance, set at the negative log10 of a p-value of 0.05 (-log10(0.05)). Longer bars in the chart suggest a more statistically significant difference between our method and the respective algorithm. For example, in the ACM dataset, the Ours method shows the most statistically significant difference when compared to AE.

To evaluate the performance of Tri-GFN across multiple datasets, we conducted a series of two-sample t-tests for comparative analysis. Tri-GFN was compared against several state-of-the-art algorithms, including KMEANS, AE, IDEC, DEC, GAE, VGAE, DAEGC, ARGA, DDGAE, DCRN, TDCN, SDCN, and MBN. We selected ACC, NMI, ARI, and F1 as performance metrics and conducted evaluations on various datasets such as ACM, Cora, DBLP, Citeseer, HHAR, Reuters, and USPS. Significance analysis results are shown in Figure 9.

### Statistical Methods

For each dataset, we calculated the t-statistic and p-value between our method and each comparison algorithm based on the four metrics. Fisher's method [51] was employed to combine individual p-values from multiple independent tests conducted across different datasets. The test statistic was computed using the formula $\chi^2 = -2 \sum_{i=1}^{k} \log(p_i)$, where $pi$ represents the p-value of the $i$-th test. The combined p-values were then transformed into their negative logarithm form for visual representation of their significance levels. Results with p-values below 0.05 were considered statistically significant.

### Detailed Analysis

On the ACM dataset, our method demonstrated high statistical significance compared to all comparison algorithms. Similar results were observed on the Cora, DBLP, and Citeseer datasets. The HHAR and Reuters datasets results indicated that our method maintained significant differences compared to most comparison algorithms. Notably, our comparisons with MBN showed significant differences across all datasets. Although our comparison with TDCN on the Citeseer dataset exhibited lower significance levels compared to other datasets, our method generally displayed significant differences against the comparison algorithms in most cases.

### 4.9  Analysis of Different Propagation Layers

In graph neural networks, propagation layers are crucial for updating node features and transmitting information. The number of these layers significantly affects the model's efficiency. To examine this, we conducted experiments with different numbers of propagation layers, evaluating performance using four key metrics.

As shown in Table 9, Tri-GFN-3, with three propagation layers, achieved the best performance across all metrics, suggesting that a three-layer structure offers an optimal balance. Except for the ACM dataset, Tri-GFN-4 generally performed worse than Tri-GFN-2 or Tri-GFN-3 on other datasets. On the USPS dataset, Tri-GFN-4's performance declined significantly, likely due to over-smoothing, which homogenizes node representations and reduces the model's ability to distinguish between nodes.

However, on the ACM dataset, Tri-GFN-4 performed best, likely because the additional layers allowed better integration of representational information. While more propagation layers can sometimes enhance performance, they can also lead to saturation or decline due to overfitting. Therefore, we selected Tri-GFN-3 as our final model to balance these effects.

**Table 9.** Propagation analysis results on seven datasets under four metrics.

| Dataset | Propagation | ACC | NMI | ARI | F1 |
|---|---|---|---|---|---|
| ACM | Tri-GFN-4 | **94.05** | 77.48 | **83.10** | **94.05** |
| | Tri-GFN-3 | 93.85 | 77.44 | 82.62 | 93.85 |
| | Tri-GFN-2 | 93.59 | 76.81 | 81.93 | 93.58 |
| | Tri-GFN-1 | 93.75 | 77.07 | 82.34 | 93.77 |
| DBLP | Tri-GFN-4 | 77.54 | 46.20 | 49.56 | 77.21 |
| | Tri-GFN-3 | **79.10** | **48.64** | **53.59** | **78.64** |
| | Tri-GFN-2 | 77.08 | 46.35 | 50.67 | 76.45 |
| | Tri-GFN-1 | 76.58 | 45.78 | 49.94 | 75.90 |
| Citeseer | Tri-GFN-4 | 66.19 | 39.55 | 40.75 | 59.81 |
| | Tri-GFN-3 | 71.60 | 46.05 | 47.57 | 65.89 |
| | Tri-GFN-2 | **71.63** | **46.19** | **47.59** | **65.90** |
| | Tri-GFN-1 | 71.57 | 45.94 | 47.61 | 65.94 |
| HHAR | Tri-GFN-4 | 63.91 | 72.76 | 59.68 | 51.75 |
| | Tri-GFN-3 | **85.08** | **81.71** | **74.80** | **85.69** |
| | Tri-GFN-2 | 78.94 | 75.58 | 67.33 | 79.13 |
| | Tri-GFN-1 | 75.84 | 72.33 | 64.99 | 73.90 |
| Reuters | Tri-GFN-4 | 69.69 | 49.13 | 49.94 | 64.41 |
| | Tri-GFN-3 | **81.93** | **56.26** | **64.56** | **73.92** |
| | Tri-GFN-2 | 81.79 | 54.82 | 63.91 | 72.76 |
| | Tri-GFN-1 | 80.69 | 53.09 | 61.78 | 72.13 |
| USPS | Tri-GFN-4 | 70.13 | 66.68 | 59.40 | 68.90 |
| | Tri-GFN-3 | **81.91** | 79.69 | **74.70** | **77.20** |
| | Tri-GFN-2 | 78.71 | **81.00** | 73.55 | 77.09 |
| | Tri-GFN-1 | 78.09 | 80.92 | 73.31 | 76.93 |
| Cora | Tri-GFN-4 | 47.67 | 29.35 | 27.21 | 25.40 |
| | Tri-GFN-3 | **74.04** | **54.24** | **51.81** | **69.24** |
| | Tri-GFN-2 | 73.15 | 53.06 | 50.63 | 68.81 |
| | Tri-GFN-1 | 71.97 | 52.42 | 49.50 | 66.21 |

The Clustering performance (%) of our model across its four versions.

Tri-GFN-x means there are x layers of encoder or decoder.

The **bold** values are the best results.

### 4.10 Analysis of Model Training Process

We analyzed the training process of the Tri-GFN model across seven datasets (ACM, DBLP, Citeseer, HHAR, Reuters, USPS, Cora), with a particular focus on the changes in four evaluation metrics over the iterations, as shown in Figure 10.

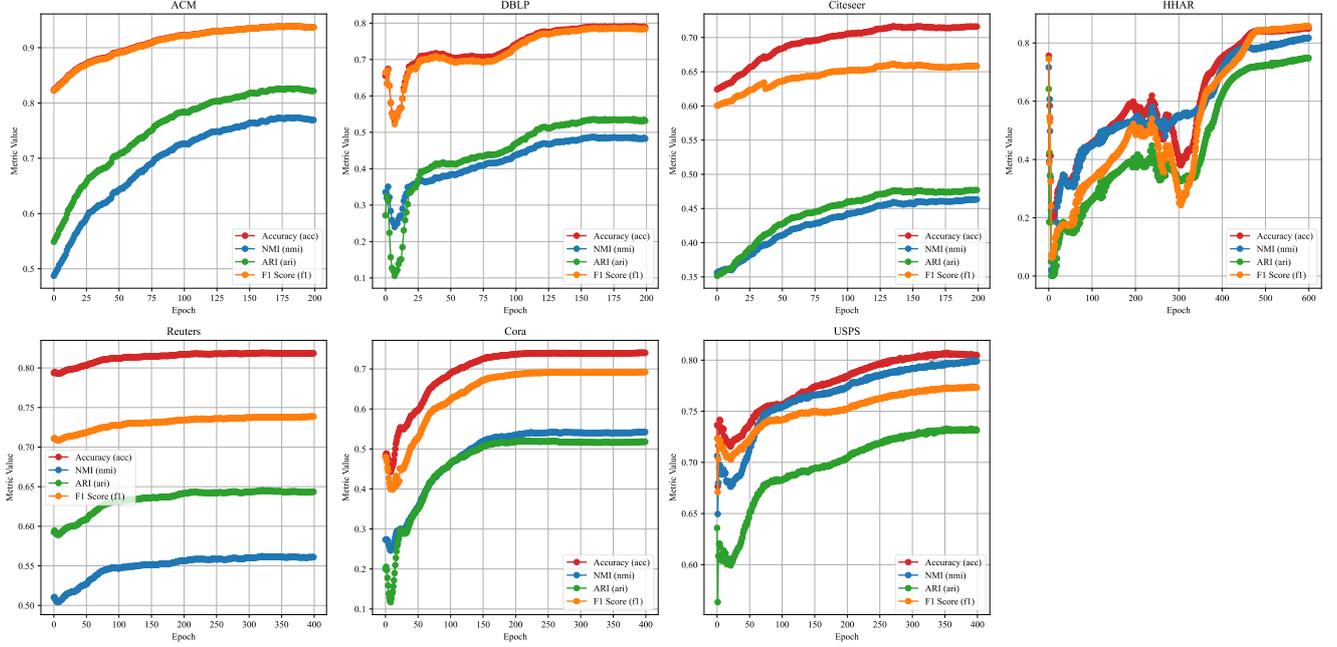

**Fig. 10** Seven scatter plots showing the changes in four evaluation metrics with an increasing number of iterations. The red line represents the ACC metric, the blue line represents the NMI metric, the green line represents the ARI metric, and the orange line represents the F1 score.

The results indicate that as the number of iterations increases, Tri-GFN's effectiveness in deep clustering of attributed graphs improves and stabilizes, validating its efficacy. For datasets like DBLP, HHAR, USPS, and Cora, the metrics initially oscillated before stabilizing, likely due to a high learning rate causing large update steps. To address this, we dynamically adjusted the learning rate, starting at a base rate and decreasing to 0.1 times every 20 epochs. While a high learning rate can cause "overshooting" and oscillations, a too-low rate can slow training and risk local optima.

## 5. Discussion

**Discussion 1: Abundance and Sensitivity of Hyperparameters**
**Impact of Hyperparameters on Model Training and Performance.**
In constructing our model, the setting of hyperparameters directly impacts the training process and final performance. In this study, we focused on the network fusion parameters $\lambda$, $\theta$, and $\gamma$, as well as the parameters $\alpha$ and $\beta$ in the total loss function. These parameters play a crucial role in adjusting our model's adaptability to different dataset characteristics. $\lambda$, $\theta$, and $\gamma$ control the weights of the GCN module, AE module, and Graph Transformer module in the enhanced module expression Eq. (15). Optimizing these parameters determines the relative importance of each module's output in the overall network expression, thereby influencing the model's ability to handle different types of structured data. The parameter $\boldsymbol{\varepsilon}$ adjusts the fusion weight between the GCN and Graph Transformer modules and the AE module at each layer, balancing the ability to extract local features and integrate global information, which is critical for capturing complex patterns and preventing overfitting. The parameters $\boldsymbol{\alpha}$ and $\boldsymbol{\beta}$ in the total loss function adjust the weights of the clustering loss $L_{clu}$ and reconstruction loss $L_{con}$, balancing the model's continuity in feature space and clustering tendency. By tuning $\boldsymbol{\alpha}$ and $\boldsymbol{\beta}$, our model can find an optimal balance between reconstruction quality and abstract feature extraction, crucial for addressing different data distribution characteristics. The learning rate is a critical hyperparameter affecting the model's training speed and stability. We set a base learning rate for each dataset to adapt to its unique data characteristics and learning difficulty. The batch size was set to 256, impacting the variance of gradient estimates and memory efficiency. A reasonable batch size helps accelerate the training process while maintaining model performance stability. Through careful hyperparameter tuning in our experiments, our model better adapted to the characteristics of different datasets, optimizing learning efficiency and generalization ability.

**Challenges of Hyperparameter Optimization.**

The selection and tuning of hyperparameters can have a significant impact on the performance of machine learning models [52], but the optimization process thus becomes particularly challenging due to the large number of hyperparameters and possible complex interactions among them. In practice, the optimal combination of hyperparameters often depends on the specific dataset characteristics and the intended model architecture, for example, in our experiments, the optimal parameter sets of $\boldsymbol{\lambda}$, $\boldsymbol{\theta}$, and $\boldsymbol{\gamma}$ are different for each dataset, and we need to manually search for the optimal parameter sets possible for each dataset. Nevertheless, by adjusting the network feature fusion weights of the three modules, we always find the optimal parameter combinations and achieve good clustering results on each dataset. This is because each module of our model exerts different effects on different dataset characteristics, and when the optimal weights for network feature fusion of the three modules are found, the three modules will exert the best results in extracting features both globally and locally. In addition, the lack of a general optimization guide makes the tuning process not only time-consuming but also inefficient. In our experiments, except for the parameter $\boldsymbol{\epsilon}$, which can be set by referring to the model setting of MBN, the rest of the parameters need to be experimented to find the optimal parameter settings for each dataset. The high-dimensionality and non-convex nature of the hyperparameter space further exacerbate this challenge, and traditional manual tuning methods are difficult to achieve near-optimal configurations in a relatively short period of time. For example, in our experiments, the initial manual tuning process to improve model performance on multiple datasets is more time-consuming, demonstrating the complexity of hyperparameter selection and the difficulty of tuning.

**Common Hyperparameter Optimization Methods.**

We tried to introduce e.g. grid search, random search [53], and Bayesian optimization [54] to try to optimize the hyperparameter problem, but both grid search and Bayesian optimization suffer from the problem of high computational cost, especially grid search, which is extremely inefficient with a large number of parameters. And although stochastic search is more efficient than grid search and can explore in a wider range of parameter spaces, in our attempts, due to the stochastic nature of stochastic search, the randomness may cause the optimization process to miss the optimal solution. In addition, the time cost required to apply these hyperparametric optimization methods in our model is no less than or even higher than manual search, due to the nature of deep clustering models, where the training process itself requires high computational resources due to its complex network structure and large number of parameters [55]. In this case, it becomes impractical to perform large-scale hyperparameter searches. For example, although Bayesian optimization can search the parameter space more efficiently in some cases, it is computationally expensive and requires retraining the model in each iteration. Therefore, we do not ultimately adopt these optimization methods to optimize the hyperparameter problem, and effective ways to solve the hyperparameter optimization problem remain to be explored.

**Sensitivity of Hyperparameters**

We conducted sensitivity analyses to understand the impact of $\lambda$ and $\theta$ on the model's performance across different datasets. We used 95% confidence intervals to explain the variability of the comprehensive metrics under different parameter values, indicating that if we repeatedly sampled from the same population and calculated the same metrics for each sample, 95% of the confidence intervals would contain the true population mean. As shown in Figure 11, the sensitivity analysis plots depict the stability and reliability of the comprehensive metrics in response to changes in $\lambda$ and $\theta$. A wide confidence interval suggests higher data variability, while a narrow interval indicates more concentrated predicted values with less uncertainty.

For different datasets, the sensitivity of $\lambda$ and $\theta$ varied significantly. On the ACM dataset, $\lambda$ peaked around 0.5, with higher sensitivity to $\theta$ in the mid-range. On the DBLP dataset, the mid-range values of $\lambda$ and $\theta$ had a significant impact on the comprehensive metrics, indicating strong sensitivity. On the Citeseer dataset, the metrics were very sensitive to $\theta$, especially at higher values, while the impact of $\lambda$ was relatively smaller. On the HHAR dataset, the metrics were significantly higher at low $\lambda$ values and mid-range $\theta$ values. On the Reuters dataset, the comprehensive metrics improved significantly at high $\lambda$ values, with lower sensitivity to $\theta$, showing a more stable trend. On the Cora dataset, the metrics were more sensitive to changes in $\lambda$, especially in the lower $\lambda$ range, while for $\theta$, the metrics were more stable in the mid-range. On the USPS dataset, the metrics showed notable sensitivity in the low to mid-range values of $\lambda$ and $\theta$.

Our sensitivity analysis of $\lambda$ and $\theta$ across datasets provided a robust basis for setting hyperparameters. By quantifying the specific impact of each parameter on model performance, we identified parameter settings that could lead to performance instability and those that enhance model stability and accuracy. Our method combined theory and practice, ensuring optimal hyperparameter settings through scientific data analysis, further enhancing the model's adaptability and robustness in various testing scenarios.

### Discussion 2: Does an Increase in the Number of Network Layers Enhance the Risk of Model Overfitting?

**The Advantages and Risks of Deep Networks.**

One of the main advantages of deep networks is their powerful feature extraction capabilities. As the number of layers increases in a neural network, each layer can capture higher-level abstract features based on the previous one. This makes deep networks particularly suitable for tasks that require understanding complex and hierarchical data structures [56]. For example, on the ACM dataset, the Tri-GFN-4 model exhibited higher performance compared to models with fewer layers, possibly because its deep structure could capture richer interactions between nodes and high-level graph structural features.

**Challenges with Increased Layer Depth.**

However, increasing the number of layers also brings significant challenges in training deep neural networks. One major issue is the vanishing or exploding gradient problem. In a multi-layer network, gradients must propagate back through each layer, potentially becoming very small (vanishing) or very large (exploding) during propagation. This phenomenon makes the network difficult to learn, as the weight updates become either very small or excessively large, leading to stalled learning or model instability (Pascanu, 2013).

**Strategies to Prevent Overfitting and Mitigate Training Difficulties.**

**Early Stopping.**

To overcome the problems of overfitting and training difficulties associated with increased layer depth, we employed early stopping and residual connections strategies [69-70]. To avoid overfitting on complex datasets such as the USPS dataset, we monitored the model's performance during pretraining and implemented early stopping when performance ceased to improve. Our pretraining strategy involved stopping training after 50 epochs for all datasets. This approach effectively prevented additional training cycles from introducing overfitting, ensuring optimal generalization performance on unseen data.

**Residual Connections in the Three-Channel Network.**

In deep neural networks, particularly when constructing deep graph neural networks, residual connections are widely regarded as an effective method to address gradient vanishing and overfitting issues during the training of deep networks. Here is how residual connections were applied in our model and the potential benefits they might bring.

After each GCN layer, a residual connection can be added, where the input to the graph convolutional layer is directly added to its output. For instance, for the first layer, the equation (6) remains unchanged, i.e., $\mathbf{Z}_{GCN}^{(1)} = \sigma \left( \mathbf{EXW}_e^{(1)} \right)$. From the second layer onwards, we can introduce residual connections using the initial input $\mathbf{X}$ or the output of the previous layer $\mathbf{Z}_{GCN}^{(\ell)}$. For the second layer, the equation can be expressed as $\mathbf{Z}_{GCN}^{(2)} = \sigma \left( \mathbf{EZ}_{GCN}^{(1)} \mathbf{W}_e^{(1)} + \mathbf{Z}_{GCN}^{(1)} \right)$. For the $\ell + 1$ layer, the update formula with residual connections can be generalized as $\mathbf{Z}_{GCN}^{(\ell+1)} = \sigma \left( \mathbf{EZ}_{GCN}^{(\ell)} \mathbf{W}_e^{(\ell)} + \mathbf{Z}_{GCN}^{(\ell)} \right)$. We apply the same operation to each AE and Graph Transformer layer. We conducted experiments with the Tri-GFN-3 model, incorporating residual connections on the ACM, DBLP, and HHAR datasets. We compared the performance of the Resi-Tri-GFN model with residual connections and the Tri-GFN-4 model on the ACM, DBLP, and HHAR datasets, as shown in Table 10. The overall results across datasets indicate that the introduction of residual connections has dataset-dependent effects on model performance. In the DBLP dataset, residual connections slightly improved performance, whereas, in the ACM and HHAR datasets, the effects were not significant. The models with residual connections performed comparably to the original models on some metrics but slightly worse on others. Therefore, we did not incorporate residual connections into our Tri-GFN-4 model to address these issues. However, further exploration and optimization of residual connection strategies are needed to better adapt to the complexities of different datasets and network structures.

**Table 10.** Comparison table of performance metrics for different datasets and propagation methods.

| Dataset | Propagation | ACC | NMI | ARI | F1 |
|---------|-------------|-----|-----|-----|-----|
| ACM | Resi-Tri-GFN | 93.69 | 76.63 | 82.14 | 93.67 |
|     | Tri-GFN-4 | **94.05** | **77.48** | **83.10** | **94.05** |
| DBLP | Resi-Tri-GFN | **77.64** | **46.80** | **51.00** | 77.15 |
|      | Tri-GFN-4 | 77.54 | 46.20 | 49.56 | **77.21** |
| HHAR | Resi-Tri-GFN | 63.78 | 72.16 | 59.29 | 51.44 |
|      | Tri-GFN-4 | **63.91** | **72.76** | **59.68** | **51.75** |

The Clustering performance (%) of our model across its two versions.

Tri-GFN-4 means there are 4 layers of encoder or decoder.

The **bold** values are the best results.

**Further Integration of Theory and Practice.** We hope to more closely integrate theoretical research and practical applications by experimentally validating theoretical models' practical effectiveness and adjusting theoretical models based on feedback from actual applications.

By addressing these areas, future research can enhance the robustness and adaptability of deep learning models, making them more effective in a broader range of applications.

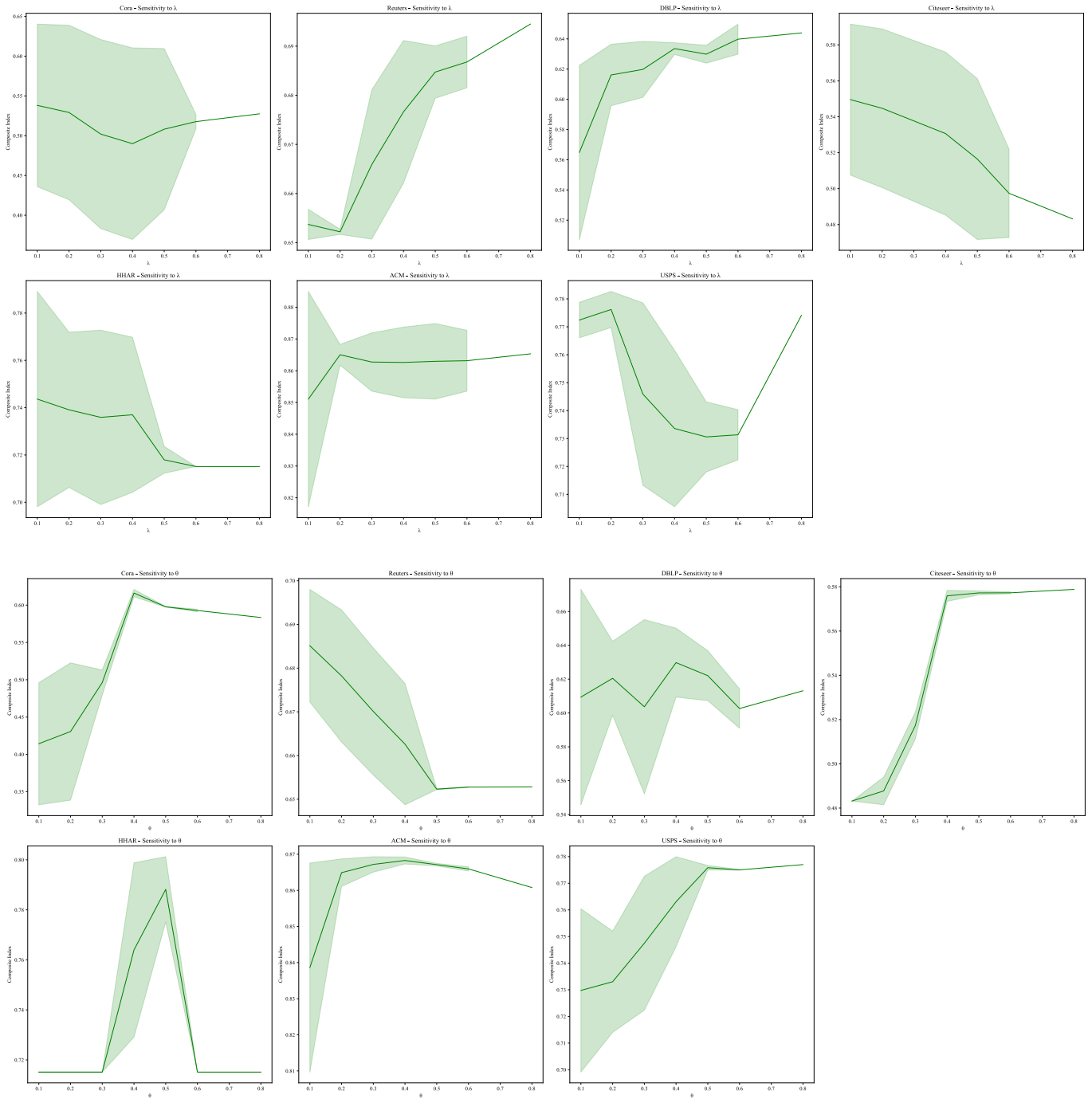

**Fig. 11** The above figure presents the sensitivity analysis of the comprehensive metric relative to parameters $\lambda$ and $\theta$ across seven datasets. Each dataset is represented by two graphs: one illustrating the sensitivity to $\lambda$ and the other illustrating the sensitivity to $\theta$.

**Conclusion and Future Work.**

We systematically explored the impact of hyperparameters on the training and performance of deep learning models and validated the effectiveness of various hyperparameter adjustment strategies through experiments. Our detailed study on hyperparameters $\lambda$, $\theta$, $\gamma$, $\alpha$, and $\beta$ revealed their critical roles in model training and demonstrated how they influence the model's adaptability and performance across different datasets. Additionally, by introducing early stopping, we effectively mitigated issues related to overfitting and training difficulties, enhancing the model's generalization capabilities. However, the problem of performance degradation caused by an excessive number of propagation layers in the network could not be resolved through early stopping or residual connections. Thus, further research is required to address the performance decline associated with using deep networks.

Despite the achievements of this study, there remain many challenges and opportunities in hyperparameter optimization. Future research can delve into the following areas:

**Automated Hyperparameter Optimization Methods.** With the continuous enhancement of computational resources and advancements in algorithms, exploring more efficient automated hyperparameter optimization techniques will be a key focus of future research. Methods based on Bayesian optimization and reinforcement learning may provide more accurate and efficient ways to find optimal hyperparameter configurations.

**Simplification and Regularization of Deep Models.** To reduce the risk of overfitting and improve model interpretability, researching more effective model simplification and regularization techniques will be of great significance.

**Cross-Domain and Cross-Task Hyperparameter Generalization.** Hyperparameter settings often depend on specific datasets and tasks. Future work can explore the generalizability of hyperparameters, aiming to discover a set of configurations that perform excellently across multiple domains or tasks.

## 6. Conclusion

This study proposes an innovative Tri-GFN designed to address the limitations of existing graph clustering methods. Tri-GFN integrates GCN, AE, and Graph Transformer modules, efficiently merging node attributes and graph structure information to enhance clustering performance. Experimental results demonstrate that Tri-GFN significantly outperforms thirteen comparative algorithms on seven public datasets, particularly excelling in handling complex relationships and high-dimensional data. Specifically, the AE module extracts node attribute representations through fully connected linear layers, while the GCN module learns deep relationships and attributes of nodes via graph convolution operations. The Graph Transformer module captures complex relationships between nodes using a self-attention mechanism. The tri-channel enhancement module further improves representation quality through inter-layer information propagation and structural information integration. The self-supervised module optimizes the clustering process by calculating soft clustering assignments and generating target distributions. The comprehensive optimization function combines reconstruction loss and clustering loss, balancing the contributions of each module, achieving optimal graph clustering performance under self-supervised learning and multi-module integration. Despite the significant advancements, some limitations remain, such as the training efficiency on large-scale graph data and hyperparameter optimization issues, which require further investigation. Future work will focus on the research of automated hyperparameter optimization techniques to reduce the cost of manual adjustments; the simplification and regularization of the model to enhance generalization ability and training efficiency; the study of cross-domain and cross-task hyperparameter generalization to explore adaptability in different application scenarios; and further integration of theory and practice to improve feasibility and performance in real-world applications.

We believe that Tri-GFN provides an efficient and innovative solution for graph clustering tasks, advancing the development of graph clustering technology. This model shows broad application potential in fields such as social network analysis, bioinformatics, recommendation systems, and network security.

## Contribution statement
**Binxiong Li:** Conceptualization, Methodology, Software, Validation, Formal analysis, Writing-Original draft, Review & Editing. **Yuefei Wang:** Conceptualization, Methodology, Project administration, Supervision, Review & Editing, Funding acquisition. **Xu Xiang:** Methodology, Software, Validation, Investigation, Data curation, Writing-Original draft, Review & Editing. **Xue Li:** Methodology, Software, Validation, Investigation, Data curation, Writing-Original draft, Review & Editing. **Binyu Zhao:** Data curation, Validation. **Heyang Gao:** Data curation, Validation. **Qinyu Zhao:** Data curation, Visualization. **Xi Yu:** Project administration.

## Acknowledgment
The authors thank the editors and the anonymous referees for their valuable comments and efforts. This research is supported by the Sichuan Comic and Animation Research Center, Key Research Institute of Social Sciences of Sichuan Province (DM2024013)